\documentclass[10pt,journal,compsoc]{IEEEtran}
\ifCLASSOPTIONcompsoc
  \usepackage[nocompress]{cite}
\else
  \usepackage{cite}
  
\fi
\ifCLASSINFOpdf
  \usepackage[pdftex]{graphicx}
\else
  \usepackage[dvips]{graphicx}
\fi
\usepackage{hyperref}
\usepackage{booktabs}
\usepackage{multirow}
\usepackage[table,xcdraw]{xcolor}
\usepackage{subfigure}
\usepackage{enumitem}
\newsavebox{\imagebox}

\usepackage{array}
\renewcommand{\arraystretch}{1.3}

\usepackage{amsmath}
\usepackage{comment}
\usepackage[utf8]{inputenc}
\usepackage[mathlines, switch]{lineno}
\usepackage{amsfonts}
\setlist[itemize]{noitemsep, topsep=0pt}
\def\eg{\textit{e.g.}}
\def\ie{\textit{i.e.}}

\def\etal{\textit{et al.}}

\begin{document}


%
\title{Explainable Human-centered Traits from Head Motion and Facial Expression Dynamics}
\author{Surbhi~Madan,\IEEEmembership{} Monika~Gahalawat,~\IEEEmembership{Student Member, IEEE}, Tanaya~Guha,~\IEEEmembership{Member, IEEE}, Roland Goecke,~\IEEEmembership{Senior Member, IEEE} and~Ramanathan~Subramanian,~\IEEEmembership{Senior Member, IEEE}

\IEEEcompsocitemizethanks{\IEEEcompsocthanksitem Surbhi Madan is with Department of Computer Science, Indian Institute of Technology Ropar, India. 
\IEEEcompsocthanksitem Monika Gahalawat, Roland Goecke and Ramanathan Subramanian are with the Human-Centred Technology Research Centre, Faculty of Science and Technology, University
of Canberra, ACT, Australia.
\IEEEcompsocthanksitem Tanaya Guha is with the School of Computing Science, University of Glasgow, UK.  

}
}

\IEEEtitleabstractindextext{%
\begin{abstract}
   We explore the efficacy of multimodal behavioral cues for explainable prediction of \textit{personality} and \textit{interview}-specific traits. We utilize elementary head-motion units named \textit{kinemes}, atomic facial movements termed \textit{action units} and \textit{speech features} to estimate these human-centered traits. Empirical results confirm that kinemes and action units enable discovery of multiple trait-specific behaviors while also enabling explainability in support of the predictions. For fusing cues, we explore decision and feature-level fusion, and an additive attention-based fusion strategy which quantifies the relative importance of the three modalities for trait prediction. Examining various long-short term memory (LSTM) architectures for classification and regression on the MIT Interview and First Impressions Candidate Screening (FICS) datasets, we note that: (1) Multimodal approaches outperform unimodal counterparts; (2) Efficient trait predictions and plausible explanations are achieved with both unimodal and multimodal approaches, and (3) Following the \textit{thin-slice} approach, effective trait prediction is achieved even from two-second behavioral snippets. 

\end{abstract}

}

\maketitle

\IEEEdisplaynontitleabstractindextext

%
\IEEEpeerreviewmaketitle

\ifCLASSOPTIONcompsoc
\else


\fi

\section{Introduction}
Personality is a psychological construct that describes human behavior in terms of habitual and fairly stable patterns of emotions, thoughts, and attributes~\cite{junior2019first,vinciarelli2014survey}. Personality is typically characterized by the OCEAN traits typified by the big-five model~\cite{mccrae1987validation}: Openness (creative vs conservative), Conscientiousness (diligent vs disorganized), Extraversion (social vs aloof), Agreeableness (empathetic vs distant) and Neuroticism (anxious vs emotionally stable). Other popular personality models include the big-two model which categorizes these five traits into the Plasticity and Stability dimensions~\cite{digman1997higher}, and the 16 personality factors model~\cite{cattell2008sixteen}. 

Personality plays a crucial role in shaping an individual's behavioral and communication traits, and how one conducts themselves in different social situations. To this end, multimodal non-verbal cues are critical in exhibiting an individual's inter-personal skills in the context of `multimedia CVs'~\cite{Raza1987AMO, Batrinca11}. Subjective impressions of interviewee's personality traits can influence hiring decisions~\cite{VanDam03}, and even one behavioral modality can explain personality attributions~\cite{DeGroot09}. \textit{E.g.}, {Conscientiousness} characterizing diligence and honesty is reflected in an upright posture and minimal head movements, while {Neuroticism} indicating anxiety and stress is revealed through fidgeting and camera aversion in self-presentation videos~\cite{Batrinca11}.  

\begin{figure}[!htb]
      \centering
     \includegraphics[width=\linewidth]{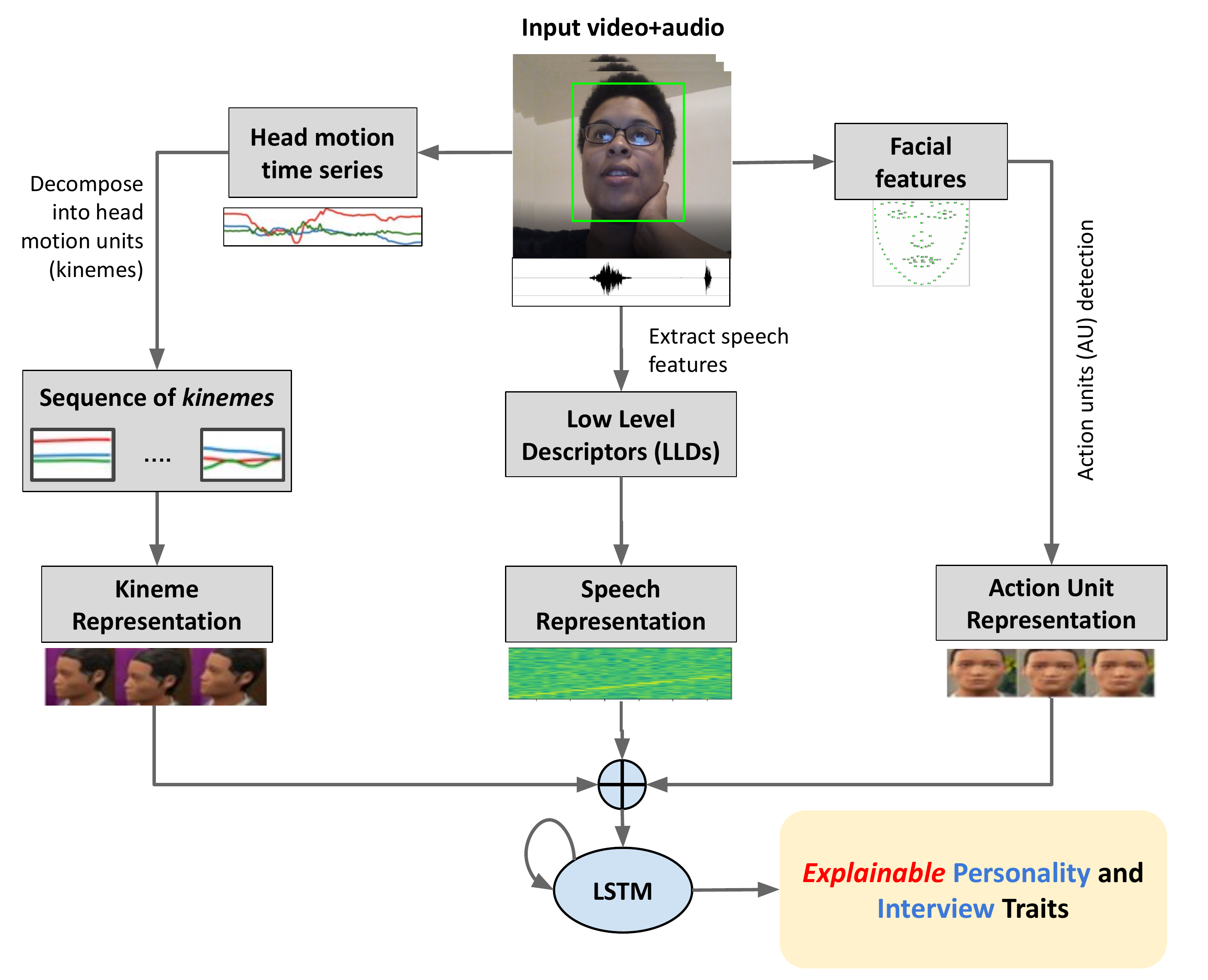}\vspace{-2mm}
    \caption{Overview of the proposed framework: Kinemes (elementary head motions), action units (atomic facial movements) and speech features employed for explainable trait prediction.} \label{fig:framework}\vspace{-4mm}
\end{figure}

This paper builds on the above findings, and explores the efficacy of multimodal behavioral cues to \textit{explainably} predict personality and job interview traits. In particular, we examine (i) elementary head motions termed \textit{kinemes}, (ii) atomic facial movements called \textit{action units} (AUs), and (iii) prosodic and acoustic speech features for traits prediction (see Fig.~\ref{fig:framework} for an overview). We first evaluate the efficacy of unimodal temporal characteristics of individual behavioral channel in predicting these traits using long-short term memory (LSTM) architectures. Next, we explore different multimodal fusion strategies (feature fusion, decision fusion, and additive soft attention) to enhance each channel's predictive power and explainability. Recent studies have already shown the effectiveness of kineme patterns for emotional trait prediction~\cite{samanta2017role, samanta2020emotion}, while acoustic features and facial expressions have been successfully employed for estimating personality attributes~\cite{junior2019first, sidorov2014automatic, kampman2018investigating} and candidate \textit{hireability} (\ie, suitability to hire/interview later)~\cite{malik2020empathetic, eddine2017personality}.  
Examining various LSTM architectures for classification and regression on the diverse FICS~\cite{escalante2020modeling} and MIT interview~\cite{naim2016automated} datasets, we make the following observations: (i) Both kinemes and AUs achieve explanative trait prediction. (ii) Multimodal approaches leverage cue-complementarity to better predict interview and personality attributes than unimodal ones. (iii) Trimodal fusion-based attention scores enable behavioral explanations, and provide insights into the relative contribution of each modality over time. (iv) Adequate predictive power is achieved even with 2 seconds-long behavioral episodes or \textit{slices}. Overall, we make the following research contributions: 
\begin{itemize}
\item Building upon our initial results~\cite{madan2021head}, we novelly employ kinemes, action units and speech features for the estimation of personality and interview traits. Given the strong correlations among personality and interview traits~\cite{Gucluturk2018,escalante2020modeling}, we show that the three behavioral modalities are both predictive and explanative of these traits. We explore distinct strategies for temporally fusing behavioral features. Fusion approaches outperform unimodal ones by a large margin owing to the complementary nature of the cues and modalities. 
\item Our experiments reveal that speech features are highly predictive of interview traits on the MIT dataset~\cite{naim2016automated}, and achieve performance comparable to kinemes and AUs for OCEAN trait prediction on the FICS dataset. 
\item Kineme and AU features enable behavioral explanations to support their predictions. We employ scores obtained from the additive attention fusion model to assess the relative importance of our three modalities per trait. 
\item We perform ablative studies presenting unimodal and multimodal results over thin-slices of varying lengths. We show that satisfactory continuous and discrete trait prediction performance can be achieved even with 2s slices, with more accurate predictions possible over longer slices in line with expectation. 
\end{itemize}

\section{Literature Review}
This section reviews research on (a) personality and interview trait prediction, and (b) multimodal behavior analytics to position our work with respect to the literature.

\subsection{Trait Prediction} 
Human thoughts, emotions and behavioral patterns are influenced by their personality, typically characterized via the OCEAN model~\cite{mccrae1987validation} characterizing human personality in terms of Openness, Conscientiousness, Extraversion, Agreeableness and Neuroticism. Various non-verbal behavioral cues such as eye movements~\cite{hoppe2018eye, rauthmann2012eyes}, head motion~\cite{jayagopi2009modeling, subramanian2013relationship}, and facial features~\cite{kampman2018investigating, guccluturk2017multimodal} have been employed for personality trait prediction. 

Numerous studies have examined the relationship between a candidate's personality traits and their job-interview performance~\cite{naim2016automated, malik2020empathetic}; For instance, Conscientiousness is positively correlated with job and organizational performance~\cite{hassan2016impact, moy2004selection}. Conscientiousness and Extraversion impact interview success~\cite{tay2006personality, barrick1991big} and job ratings~\cite{witt2002interactive}. While Mount~\etal \cite{mount1998five} observed that Emotional stability, Conscientiousness and Agreeableness are positively related to job performance, Rothmann\etal~\cite{rothmann2003big} associated Conscientiousness, Extraversion, Emotional stability and Openness with job performance and creativity. While these correlations among personality and interview traits have been discovered via statistical analyses, very few studies have explored the relationships between non-verbal behavioral cues and personality-cum-interview traits in a predictive (regression/classification) setting.  


 \textbf{Explainable trait prediction:} Despite achieving excellent performance on multiple prediction problems, deep learning models fall short in terms of explainability and interpretability due to their `black-box' nature~\cite{samek2019towards}. Recent studies alleviate this issue by interpreting the results of deep learning models, \eg, Wicaksana and Liem~\cite{wicaksana2017human} predict OCEAN personality traits explicitly focusing on human-explainable features and a transparent decision-making process. Wei \etal~\cite{wei2017deep} propose a deep bimodal regression framework, in which Convolutional Neural Networks (CNNs) are modified to aggregate descriptors for improving regression performance on apparent personality analysis. A CNN-based approach for interpretability is explored, where the authors observe a correlation between AUs and CNN-learned features \cite{ventura2017interpreting}. Another work \cite{gucluturk2017visualizing} trains a deep residual network with audiovisual descriptors for personality trait prediction, where predictions are elucidated via face image visualization and occlusion analysis. 

\subsection{Multimodal Behavior Analytics} Low-level behavioral features have been largely employed for human-centred trait prediction. \textit{E.g.}, head-motion has been modeled with descriptors such as amplitude of Fourier components~\cite{ding2018low}, Euler rotation angles and velocity. Head motion is often restricted to nods and shakes \cite{gunes2010dimensional}. Yang and Narayanan~\cite{yang2016modeling} extract arbitrary head motion patterns, which do not have a physical interpretation. Subramanian~\etal~\cite{subramanian2013relationship} predict Extraversion and Neuroticism employing positional and head pose patterns.

Audio-visual features are typically combined to achieve effective trait prediction. Low-level speech descriptors such as pitch, intensity, spectral, cepstral coefficients and pause duration are commonly used for personality \cite{an2018lexical,valente2012annotation} and affect recognition \cite{mangalam2017learning, tawari2010speech, abdel2020egyptian}. Other works use acoustic, prosodic and linguistic features for personality prediction~\cite{levitan2016identifying, kampman2018investigating}.

Many trait prediction studies focus solely on visual cues, with facial cues playing a crucial role. \textit{E.g.}, multivariate regression is employed to infer user personality impressions from Twitter profile images~\cite{dhall2016first}, while eigenfaces are combined with Support Vector Machines are used to predict if a depicted person scores above/below the median for each of the big-five traits~\cite{al2014face}. Meng \etal~\cite{meng2021factors} investigate the connection between gratification-sought (\eg, escape, fashion, entertainment) and personality traits, and find that extroverts are more active in contributing to, and participating in engaging behaviors. Short-term facial dynamics are learned from short videos via an emotion-guided, encoder-based approach for personality analysis in~\cite{song2021self}.

\subsection{Summary} 
Our literature review reveals the following research gaps: 
 \begin{itemize}
     \item[(1)] Personality and interview traits are known to be highly correlated based on statistical observations, but few works have explored learning of features that can effectively predict as well as explain these traits. 
     \item[(2)] While personality and interview traits have been predicted via machine/deep learning approaches, the majority employs statistics of low-level audiovisual features (statistics relating to head motion, eye-gaze, facial expression, speech and prosodic), which  limits {explanations} to support the predictions. While head motion patterns have been identified as critical non-verbal behavioral cues, they have not been employed for personality or interview trait prediction. We show how kineme and AU features can intuitively explain trait-specific behaviors.
     \item[(3)] Multimodal behavioral analytics have been largely restricted to feature and decision fusion, treating all behavioral channels equally. Differently, we utilize additive soft attention~\cite{sharma2018multichannel}-based fusion that learns relative contribution of each channel from data. This allows for quantifying and explaining the relative contribution of the different modalities towards the prediction result. 
 \end{itemize}


\section{Methodology} \label{features}
\subsection{Feature Extraction}
We now present feature extraction for the three employed modalities: (i) 3D head motions denoted via a sequence of kinemes, (ii) facial action units describing muscle movements, and (iii) low-level descriptors for speech representation. As in~\cite{madan2021head}, we encode these features into 2s temporal segments with a 50\% overlap to obtain feature vectors.  

\textbf{Kineme Representation:}
A compact approach to modeling head motion is by representing it in terms of a small number of fundamental and interpretable units termed \textit{kinemes} \cite{samanta2017role}; they are analogous to phonemes in human speech \cite{birdwhistell2010kinesics}. We extract the 3D Euler rotation angles \emph{pitch} ($\theta_p$), \emph{yaw} ($\theta_y$) and \emph{roll} ($\theta_r$) per frame to represent head pose using the Openface toolkit \cite{baltruvsaitis2016openface}. Head motion for a time period  $T$  can be represented as a time-series of 3D angles: $\boldsymbol{\theta} = \{\theta_p^{1:T}, \theta_y^{1:T}, \theta_r^{1:T}\}$. 
This multivariate time-series $\boldsymbol{\theta}$ of length $T$ is divided into $l$-overlapping segments, where the $i^{th}$ segment is denoted by a vector 
$\mathbf{h}^{(i)}= [\theta_p^{i:i+\ell}, \theta_y^{i:i+\ell}, \theta_r^{i:i+\ell}]^T$. 
These overlapping segments enable shift-invariance and generate better representations of the head motion \cite{samanta2020emotion}. 

Further, we define the characterization matrix as $\mathbf{H}_{\boldsymbol\theta} = [\mathbf{h}^{(1)}, \mathbf{h}^{(2)},\cdots, \mathbf{h}^{(s)}]$ with $s$ denoting the number of segments in the training sample. All $N$ training samples are combined to form the head motion matrix $\mathbf{H} = [\mathbf{H}_{\boldsymbol\theta_1}|\mathbf{H}_{\boldsymbol\theta_2}|\cdots|\mathbf{H}_{\boldsymbol\theta_N}]$, where each column in the matrix $\mathbf{H}$ represents a single head motion time-series segment. Non-negative Matrix Factorization is performed on the matrix $\mathbf{H}$ to obtain basis and coefficient matrices $\mathbf{B}$ and $\mathbf{C}$ respectively. We then employ Gaussian Mixture modeling to cluster coefficient vectors in a low dimensional space to obtain a $k$ column matrix ${\mathbf{C}^*}$ ($k<<Ns$). The matrix ${\mathbf{C}^*}$ is transformed as $\mathbf{H}^*=\mathbf{B}\mathbf{C}^*$, to obtain  kinemes in the original space. Columns of $\mathbf{H}^*$ yield the $k$ kinemes $\{\mathcal{K}_i\}_{i=1}^K$. 

\begin{figure}[!htb]
      \centering
     \includegraphics[width=\linewidth]{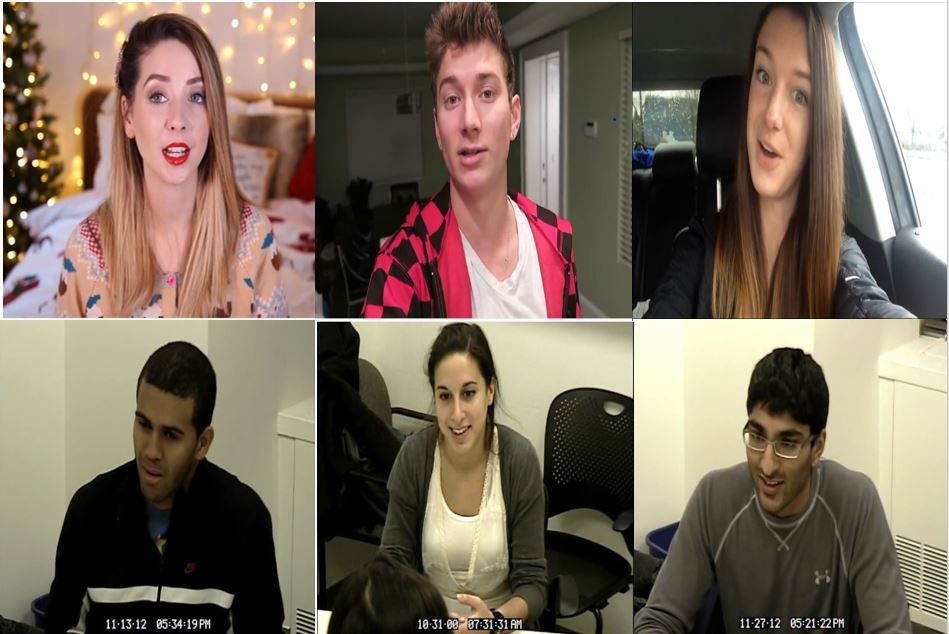}\vspace{-2mm}
    \caption{FICS (top) and MIT (bottom) exemplars.} \label{fig:dataset_frames}\vspace{1mm}
      \centering
     \includegraphics[width=\linewidth]{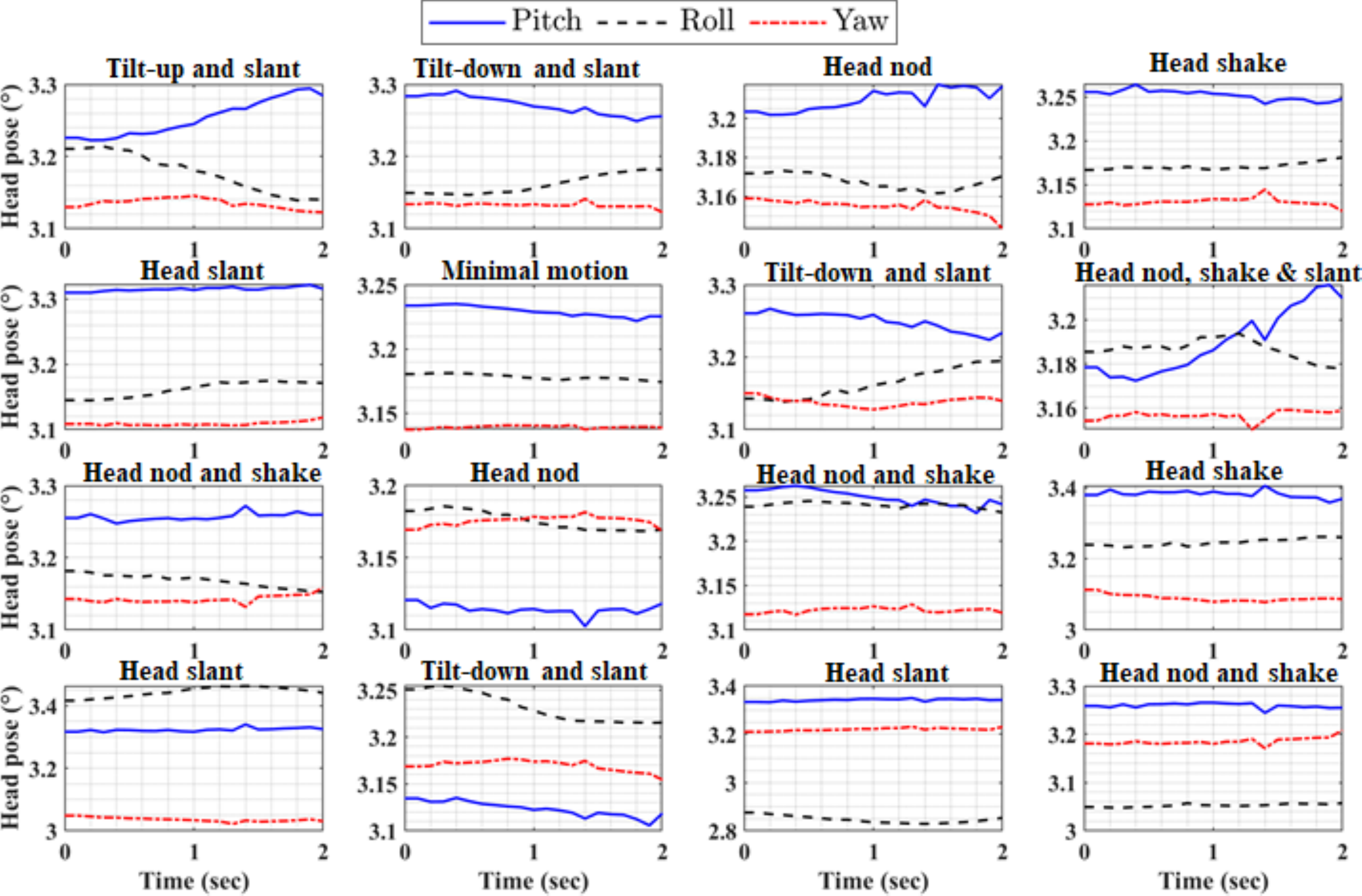}\vspace{-1mm}
    \caption{Plots of 16 kinemes extracted for the FICS dataset following raster ordering (left to right, top to bottom.)}\vspace{1mm} \label{fig:selectKinemes_FICS}
      \centering
     \includegraphics[width=\linewidth,height=2.3cm]{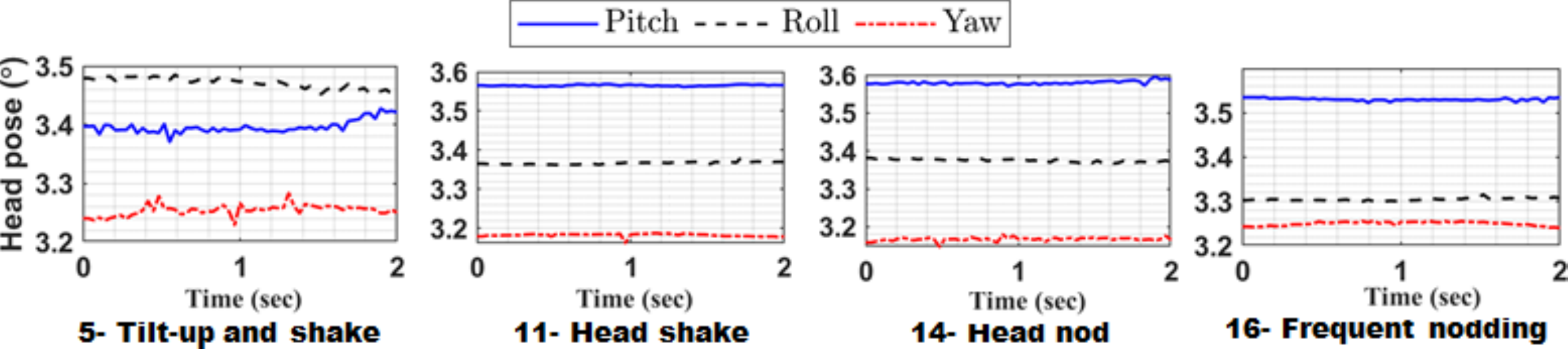}\vspace{-1mm}
    \caption{Selected kineme plots for the MIT dataset.} \label{fig:selectKinemes_MIT} \vspace{1mm}

     \includegraphics[width=\linewidth]{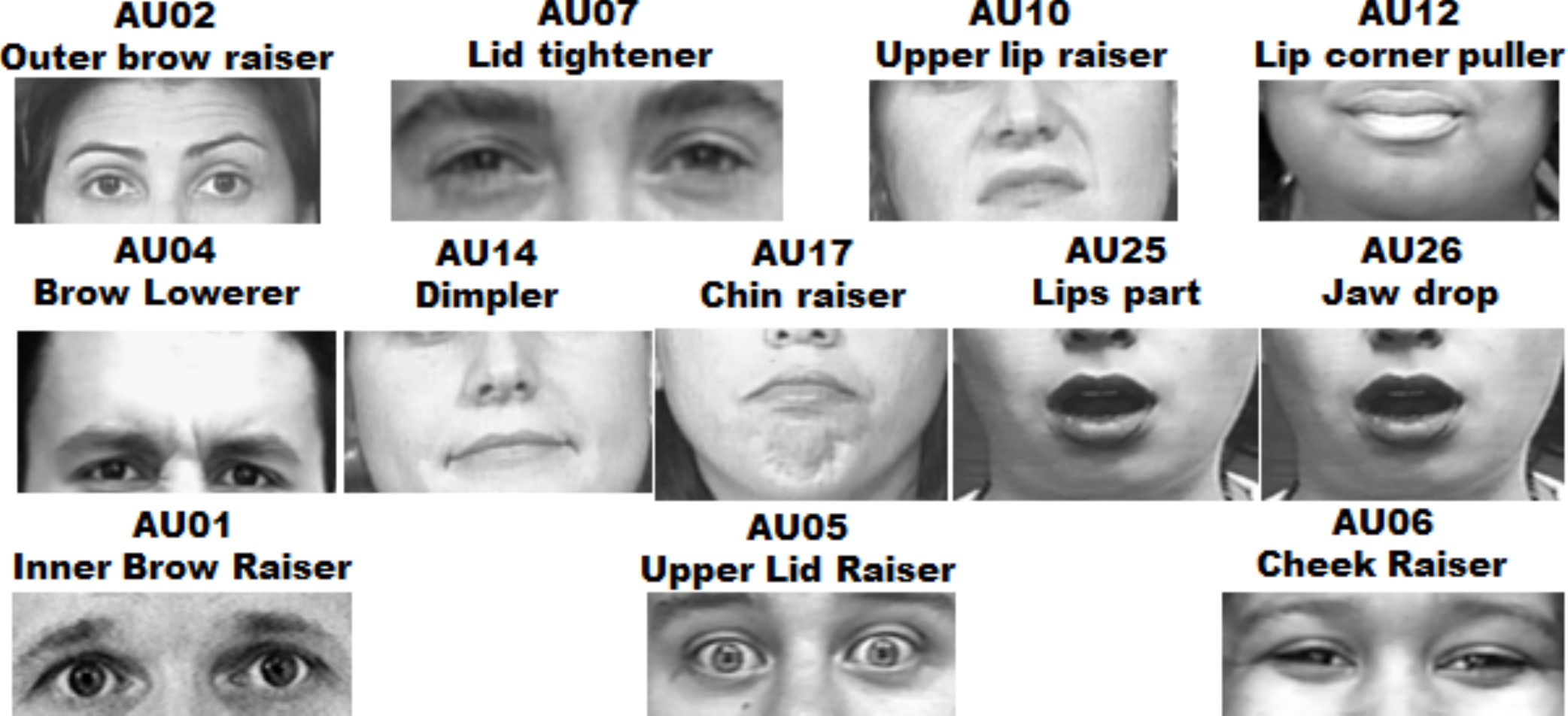}\vspace{-2mm}
    \caption{Common AUs in the FICS and MIT datasets.} \label{fig:selectAUs}\vspace{-.4cm}
\end{figure}

On learning the kineme representation, any head motion time-series is represented via $\mathcal{K}$ by mapping each time series segment to an individual kineme. To obtain the corresponding kineme, we compute the characterization matrix $\mathbf{h}^{(i)}$ for the $i^{th}$ segment. Lastly, we project $\mathbf{h}^{(i)}$ onto the learned subspace spanned by $\mathbf{B}$ to get $\mathbf{c}^{(i)}$:
\begin{equation*}
    \hat{\mathbf{c}} = \underset{\mathbf{c}^{(i)} \geq 0}{\text{arg min}} \lVert{\mathbf{h}^{(i)} - \mathbf{B}\mathbf{c}^{(i)}}\rVert_F^2
\end{equation*}
We maximize the posterior probability $P({\mathcal{K}_i}|\hat{\mathbf{c}})$ to associate the  $i^{th}$ segment to its corresponding kineme $\mathcal{K}_i$. Thus, we can map a head motion time-series to a kineme sequence. Selected kinemes are extracted from the MIT and FICS datasets are visualized in Figs.~\ref{fig:selectKinemes_FICS} and~\ref{fig:selectKinemes_MIT}.

\vspace{2mm}
\textbf{Action Unit Detection:}
We extract 17 facial action units (AUs) per video frame using Openface. These 17 AUs are described in terms of a value specifying the visibility of an AU, and an intensity score representing AU sharpness on a 5-point scale (minimal to maximal). We employ mean intensity as a threshold to identify the dominant AUs over all 2s frames with 1s overlap as above. We present common AUs from the two datasets in Fig.~\ref{fig:selectAUs}.

\vspace{2mm}
\textbf{Speech Feature Extraction:}
We extracted low-level audio descriptors (LLDs) via the Librosa library ~\cite{mcfee2015librosa} following the Interspeech2009 emotion challenge \cite{schuller2009interspeech}: Fundamental frequency (F0), voice probability, zero-crossing rate (ZCR) and Mel-frequency cepstral coefficients (MFCCs). A local feature vector is created by extracting the LLDs over a sliding window of 93ms with an overlap of 23ms over the entire video duration. These local features are averaged and concatenated to obtain a 23-dimensional feature vector for each 2s segment. For each dataset, these features are normalized to have zero mean and unit variance.   

\begin{figure*}[!htbp]
      \centering
     \includegraphics[width=\linewidth]{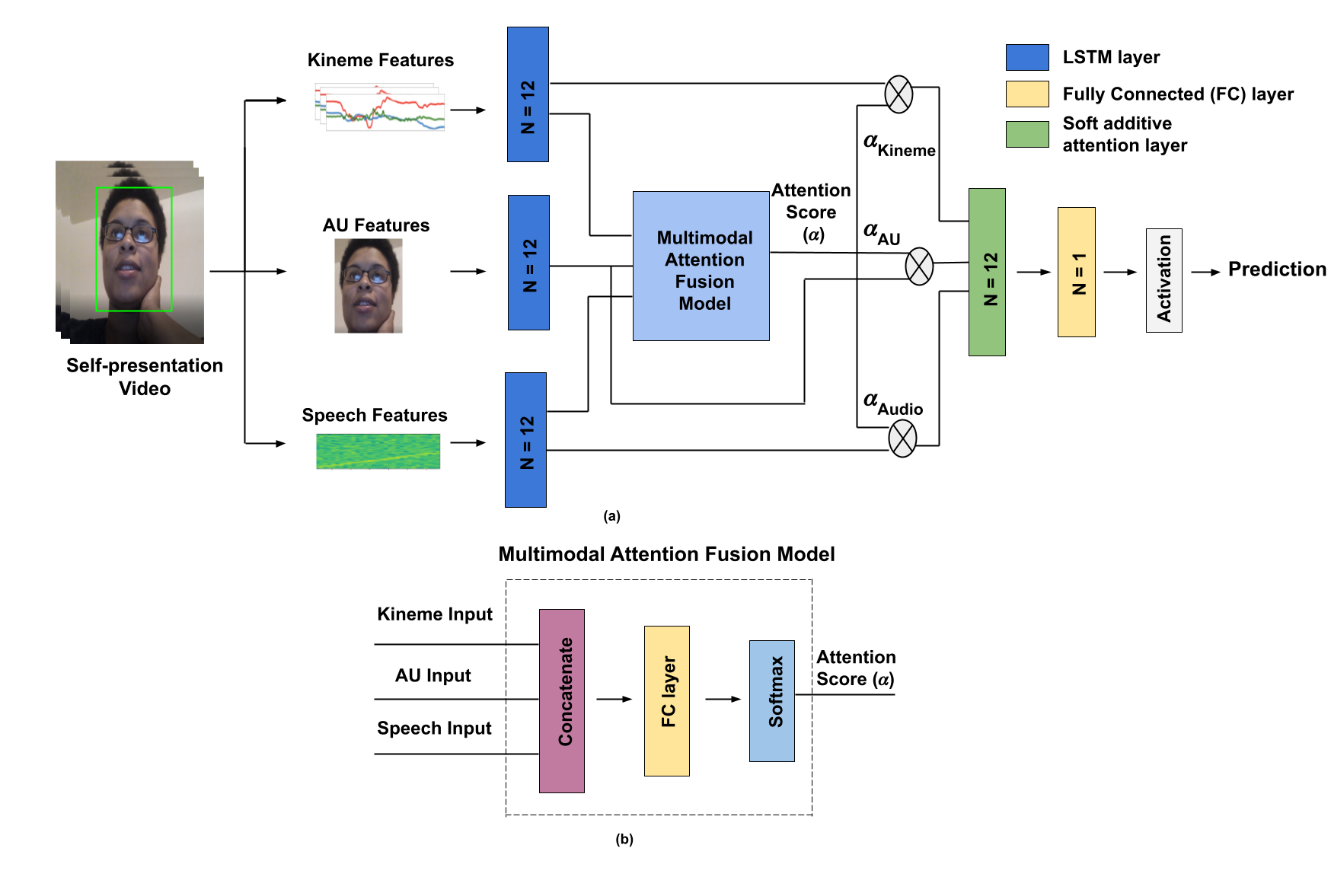}\vspace{-8mm}
    \caption{(a) Additive attention fusion architecture overview, and (b) Attention score computation process (FC layer comprises twelve neurons). $N$ denotes the number of neurons per layer. Linear/sigmoid activation is applied on the dense layer output for regression/classification.} \label{fig:LSTM_attention_arch}\vspace{-4mm}
\end{figure*}

\subsection{Models}

\textbf{Long short-term memory (LSTM)} models for regression and classification: We trained LSTMs with the kineme (\textbf{LSTM Kin}), AU (\textbf{LSTM AU}) and speech sequences (\textbf{LSTM Aud}). We also performed bimodal feature fusion (\textbf{FF}) and decision fusion (\textbf{DF}) with all combinations (\textbf{LSTM Kin+AU}, \textbf{LSTM Kin+Aud} and \textbf{LSTM AU+Aud}), and trimodal LSTM fusion (\textbf{LSTM Kin+AU+Aud}). The kineme sequences are one-hot encoded, where the kineme denoting a given time-window is coded to 1 and the rest to 0. AU sequences are encoded by setting the dominant AUs to 1 and rest to 0 for the time-window, creating a binary 17-element AU vector. Speech sequences are created by $z$-normalizing LLDs averaged over the time-window. For a behavioral slice involving $L$ time windows with $N$ training samples, the kineme, AU and speech features form 3D matrices of size 16$\times N \times L$,  17 $\times N \times L$, and 23 $\times N \times L$ respectively. 

\begin{figure}[!hb]
      \centering
     \includegraphics[width=\linewidth]{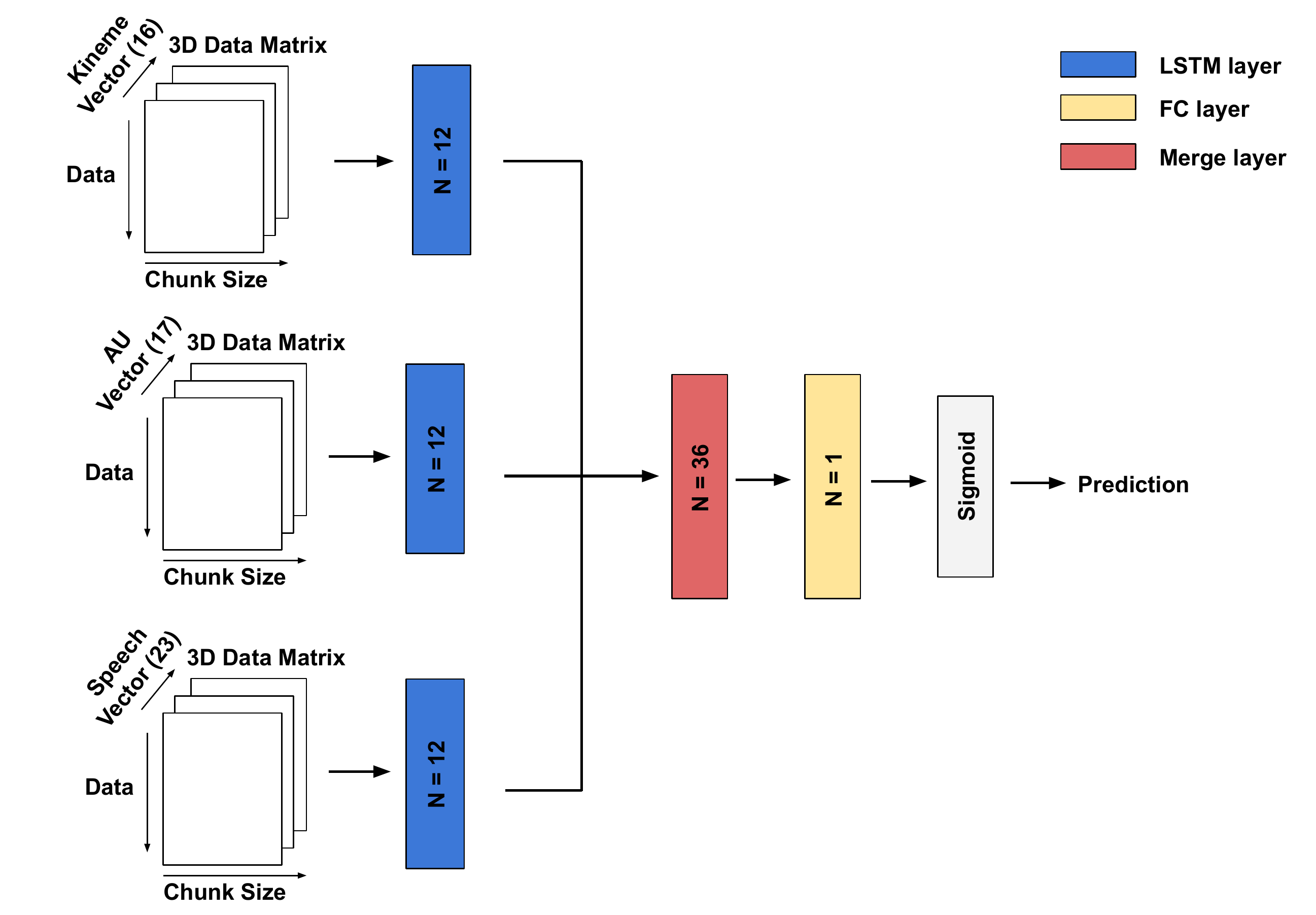}\vspace{-3mm}
    \caption{Trimodal feature fusion architecture. Linear activation is applied on the dense layer output for regression. $N$ denotes the number of neurons per layer.  The dense layer output involves linear activation and 32 neurons in the LSTM layer for regression model.} \label{fig:LSTM_decision_arch}\vspace{-2mm}
\end{figure}

\noindent \textbf{Unimodal and feature fusion (FF):} A single hidden LSTM layer is employed for unimodal prediction followed by a dense layer involving one neuron with sigmoidal/linear activation for classification/regression. For bimodal and trimodal feature fusion, unimodal descriptors are fused by applying a single LSTM layer to each feature. The subsequent outputs are merged followed by a dense layer comprising a single neuron as above (see Fig.~\ref{fig:LSTM_decision_arch}). The hyperparameters such as number of neurons, activation function and dropout rate are tuned via the validation set. An Adam optimizer is utilized for training with learning rate of 0.01. We employ binary cross entropy and mean absolute error as loss functions for classification and regression respectively. 

\noindent \textbf{Attention fusion (LSTM AF):} 
To achieve multimodal explanations, we employ attention-based trimodal fusion as in~\cite{sharma2018multichannel} to assign importance weights to the three modalities at each time window (Fig.~\ref{fig:LSTM_attention_arch}). Dense layers are employed for each cue in~\cite{sharma2018multichannel}, while we use one LSTM layer per modality to quantify an \textit{importance weight}. Also, while we compute weights based on softmax scores generated per time step,~\cite{sharma2018multichannel} focuses only on the channel with maximum attention weight discarding others.
As in Fig.~\ref{fig:LSTM_attention_arch}(a), an LSTM layer is employed for each modality to learn temporal dynamics, resulting in a fixed-length feature vector per modality. Unimodal descriptors are concatenated and passed through a fully connected layer, and a softmax layer composed of three neurons (Fig.~\ref{fig:LSTM_attention_arch}(b)). Attention scores generated via the softmax layer are deemed as the relative contribution of each modality per time window.
Layer normalization is applied over each unimodal feature vector. To fuse normalized features, we employ an additive layer to sum the weighted unimodal features. This is followed by a dense layer comprising a single neuron with sigmoidal/linear activation for classification/regression. We aggregate weights to compute modality contributions over behavioral slices spanning multiple time windows. 


\noindent \textbf{Decision fusion (DF):} We adopt the fusion weight estimation approach~\cite{koelstra2013fusion} outlined below. Assuming the unimodal classifier/regressor scores are  $p_{1}$ and $p_{2}$ for the bimodal fusion, the test sample score is defined as $\alpha p_{1} + (1-\alpha) p_{2}, \alpha \in [0,1]$. We perform grid search with a step-size of 0.05 to identify the optimal $\alpha^*$ maximizing F1-score and Pearson correlation coefficient (PCC) respectively for classification and regression (the same is extended to trimodal fusion). 

\section{Experimental Results}
\subsection{Datasets} \label{datasets}
The \textbf{FICS} dataset~\cite{escalante2020modeling} contains 10K self-presentation snippets derived from YouTube videos of people talking into the camera. Averaging 15s in length, these videos are split into a 3:1:1 proportion for train, validation and test. All videos are annotated with OCEAN trait scores with `N' scores denoting emotional stability instead of Neuroticism. This \textbf{MIT} dataset~\cite{naim2016automated} comprises audio-visual recordings of 138 mock job interviews with 69 undergraduate students, with videos being 4.7 minutes long on average. All videos are annotated with 16 interviewee-specific traits. We focus on the following traits: recommended hiring score (RH) denoting the candidate's hireability, level of excitement (Ex), friendliness (Fr) and eye-contact (EC). We also examine the Overall (Ov) interview score in prediction experiments. Examples from the two datasets are presented in Figure~\ref{fig:dataset_frames}.

\subsection{Quantitative Experiments}
\textbf{Prediction Settings:} We consider both continuous and discrete prediction of personality and interview traits. For (binary) classification, we dichotomize trait scores by thresholding them at their median value. Tables~\ref{tab:MIT_reg} and~\ref{tab:FICS_reg} present regression results, while Tables~\ref{tab:MIT_cls} and~\ref{tab:FICS_cls} present classification results. Our models are fine-tuned on the FICS dataset via the pre-defined validation set, while hyperparameter tuning is achieved via 10-fold cross-validation (cv) on the smaller MIT Interview dataset. Results reported on the MIT dataset are $\mu \pm \sigma$ statistics noted over 50 runs (5 repeated runs of 10-fold cv). Early stopping with a patience value of 4 epochs is employed to prevent model degradation.

\begin{figure*}[!htbp]
\centering
    \includegraphics[width=0.48\linewidth]{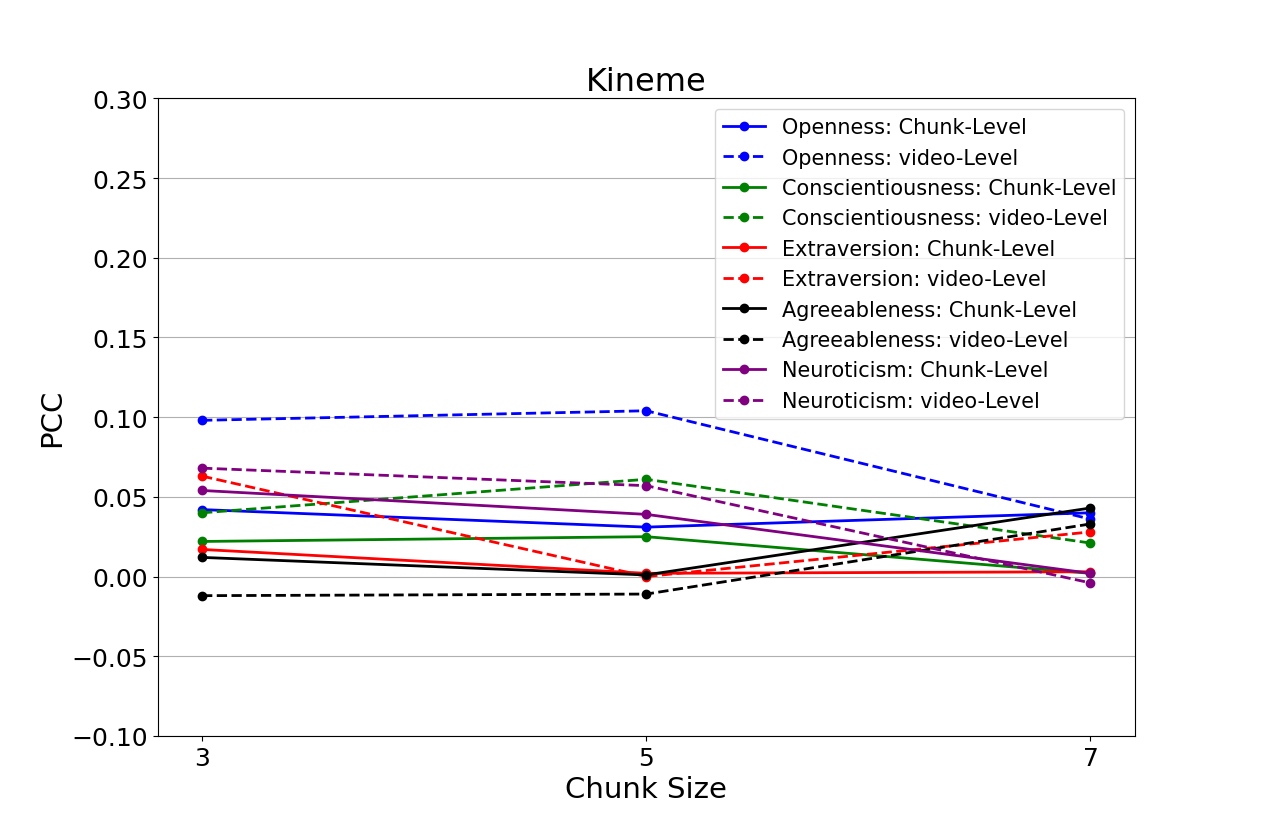}\hspace{0.05cm}\includegraphics[width=0.48\linewidth]{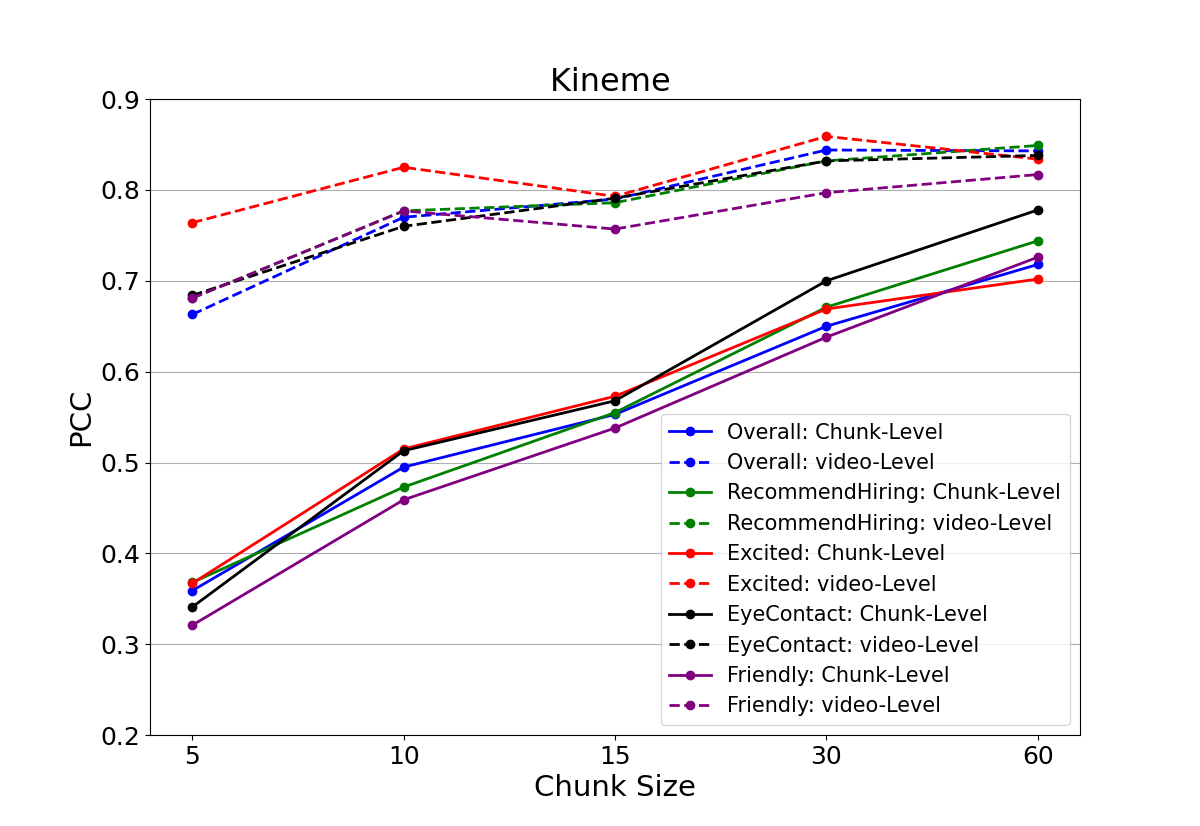}
        \vspace{-2mm}\caption{Chunk vs video-level predictions with kinemes for FICS (left) and MIT (right). dataset.}\label{fig:Chunk_vs_Vid_kineme}
    \vspace{.5mm}
        \centering
        \includegraphics[width=0.5\linewidth,height=5cm]{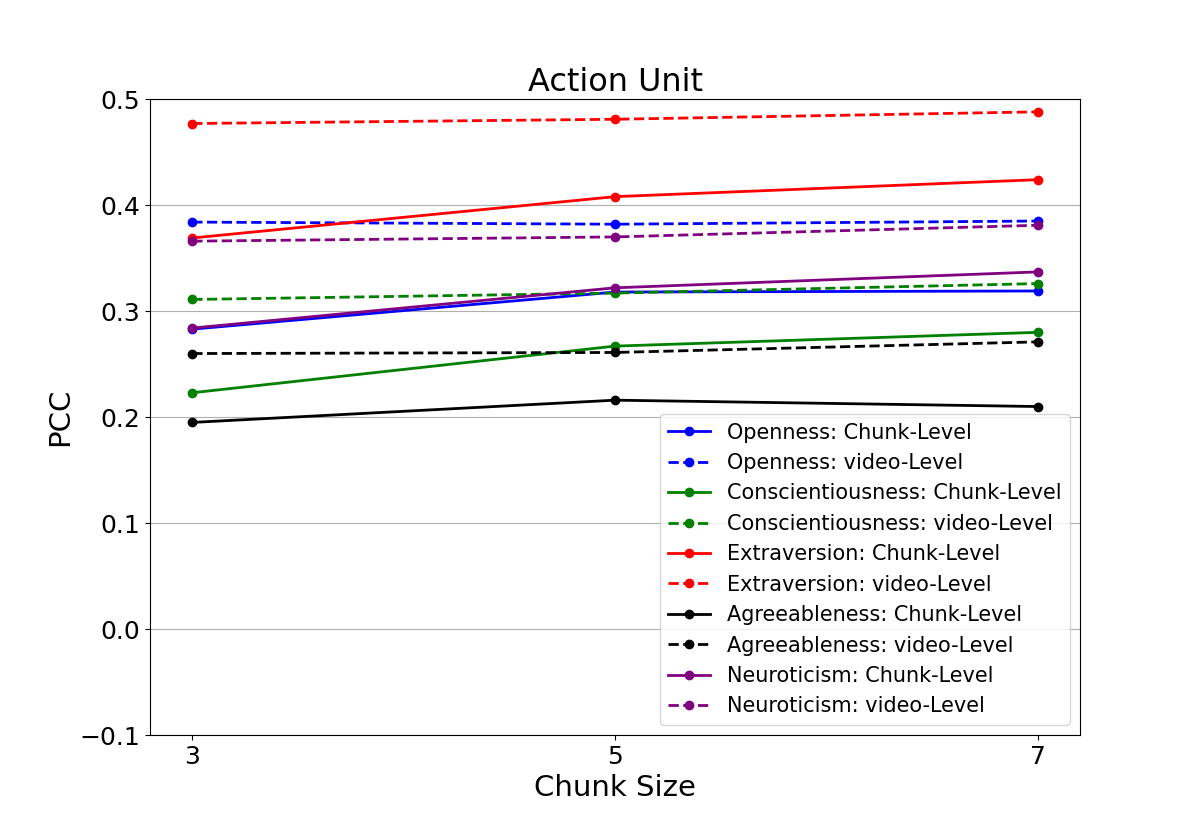}\hspace{0.05cm}\includegraphics[width=0.48\linewidth,height=5cm]{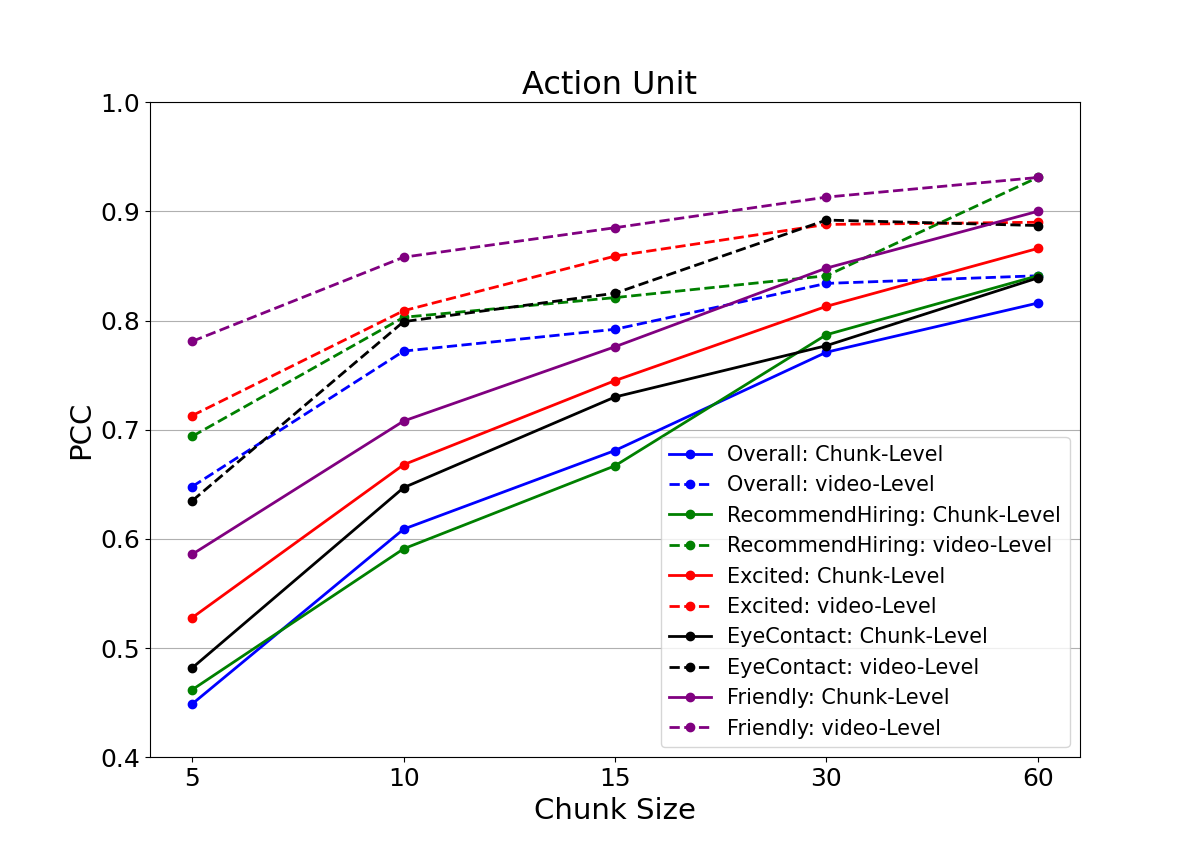}
        \vspace{-2mm}\caption{Chunk vs video-level predictions with AUs for FICS (left) and MIT (right). dataset.}\label{fig:Chunk_vs_Vid_AU}\vspace{.5mm}
        \centering
        \includegraphics[width=0.48\linewidth,height=5cm]{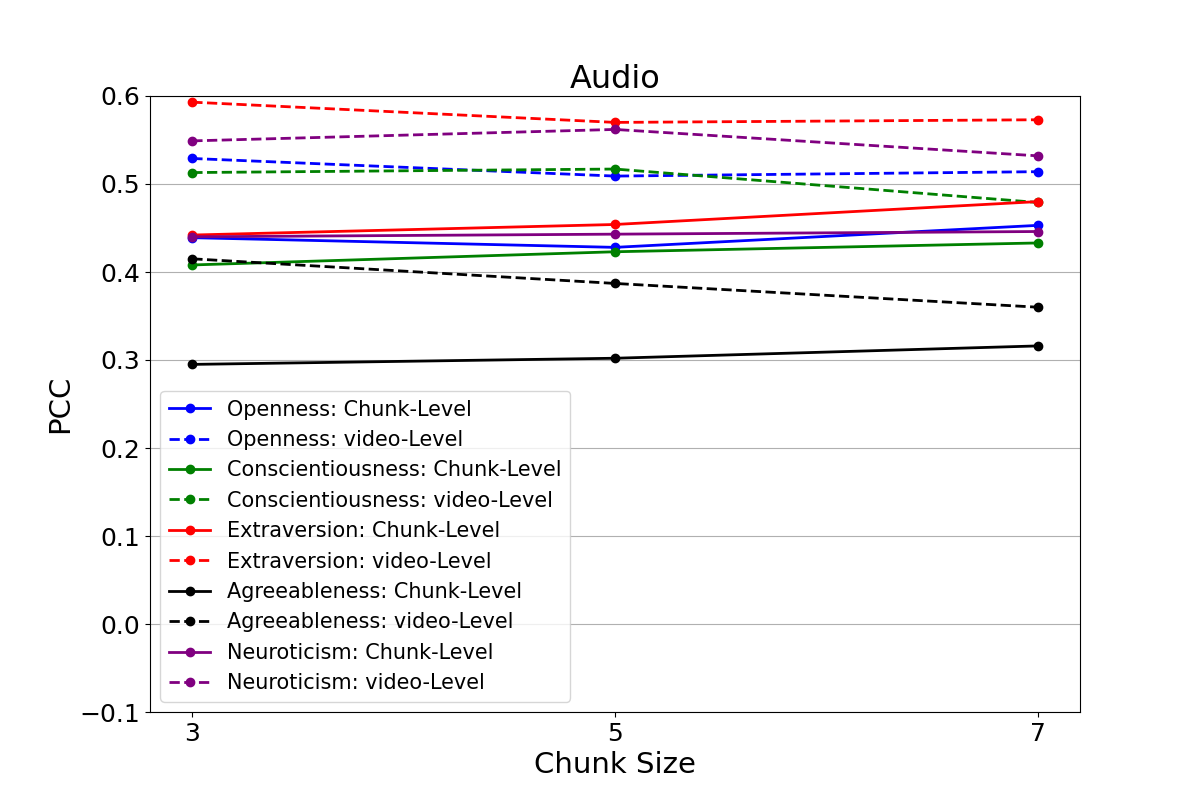}\hspace{0.05cm}\includegraphics[width=0.48\linewidth,height=5cm]{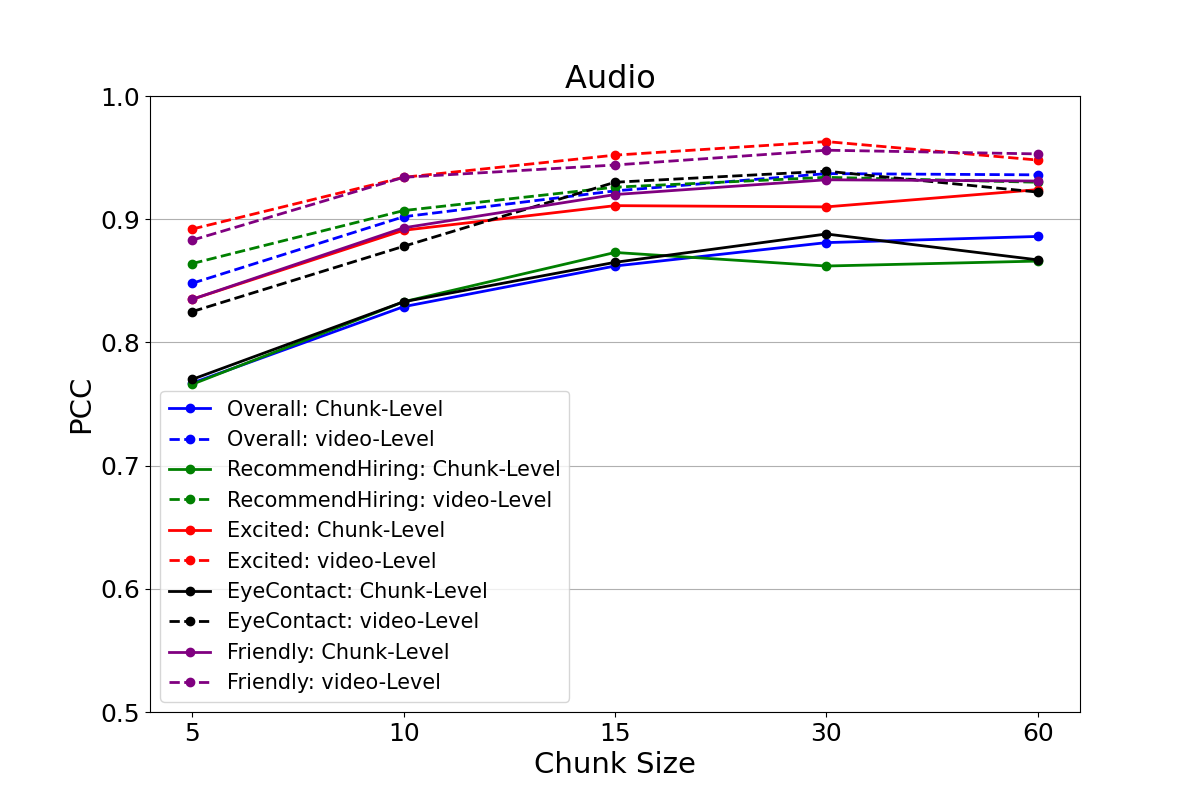}
        \vspace{-2mm}\caption{Chunk vs video-level predictions with speech features for FICS (left) and MIT (right). dataset.}\label{fig:Chunk_vs_Vid_Audio}\vspace{-3mm}
\end{figure*}

\noindent \textbf{Chunk vs video-level prediction:} To examine trait prediction over tiny behavioral episodes (or slices), we segment the original videos into smaller chunks of 2-7s for FICS, and 2-60s for the MIT dataset. All video chunks are assigned the source video label. We then compute metrics over a) all chunks (chunk-level performance), and b) over all videos by assigning the majority label/mean value over all chunks (video-level performance) for classification/regression.  A comparison of chunk vs video-level predictions for the three modalities is presented in Figs.~\ref{fig:Chunk_vs_Vid_kineme}- \ref{fig:Chunk_vs_Vid_Audio} 
\\
\textbf{Performance Metrics:}
Due to the imbalanced class distribution in classification, we use two metrics: Accuracy (Acc) and F1-Score. For regression, accuracy (Acc) defined as 1-MAE (Mean Absolute Error)~\cite{guccluturk2017multimodal} and PCC (Pearson Correlation Coefficient) are considered. 

\subsection{Results and Discussion}\label{RandD}

\begin{table*}[!htbp]
    \centering
		\fontsize{7}{7}\selectfont
		\renewcommand{\arraystretch}{1.5}
    \caption{Unimodal and multimodal regression results on the MIT dataset. Accuracy and PCC values are tabulated, with highest PCC achieved per trait denoted in \textbf{bold}.} \vspace{-2mm}
    \begin{tabular}{|l|cc|cc|cc|cc|cc|cc|}
    \hline
     \bf  & \multicolumn{6}{c|}{\textbf{Unimodal}} & \multicolumn{6}{c|}{\textbf{Trimodal}} \\ \cline{2-13}
     \bf Trait & \multicolumn{2}{c|}{\textbf{LSTM Kin}}  & \multicolumn{2}{c|}{\textbf{LSTM AU}} & \multicolumn{2}{c|}{\textbf{LSTM Audio}} & \multicolumn{2}{c|}{\textbf{LSTM FF}}  & \multicolumn{2}{c|}{\textbf{LSTM DF}}  & \multicolumn{2}{c|}{\textbf{LSTM AF}}\\ 
		& \multicolumn{1}{c}{\textbf{Acc}} &  \multicolumn{1}{c|}{\textbf{PCC}}  & \multicolumn{1}{c}{\textbf{Acc}} & \multicolumn{1}{c|}{\textbf{PCC}} & \multicolumn{1}{c}{\textbf{Acc}} & \multicolumn{1}{c|}{\textbf{PCC}} & \multicolumn{1}{c}{\textbf{Acc}} & \multicolumn{1}{c|}{\textbf{PCC}} & \multicolumn{1}{c}{\textbf{Acc}} & \multicolumn{1}{c|}{\textbf{PCC}} & \multicolumn{1}{c}{\textbf{Acc}} & \multicolumn{1}{c|}{\textbf{PCC}} \\
     \hline
      \textbf{Ov} & 0.93$\pm$0.04& 0.84$\pm$0.26& 0.93$\pm$0.04& 0.84$\pm$0.26& 0.96$\pm$0.03& 0.94$\pm$0.10& 0.97$\pm$0.03& 0.96$\pm$0.08& 0.97$\pm$0.01& \textbf{0.97$\pm$0.04}& 0.98$\pm$0.03& 0.95$\pm$0.16\\
			\textbf{RH}    &  0.95$\pm$0.03& 0.93$\pm$0.10& 0.95$\pm$0.03& 0.93$\pm$0.10& 0.96$\pm$0.03& 0.93$\pm$0.09& 0.97$\pm$0.03& 0.96$\pm$0.08& 0.98$\pm$0.01& \textbf{0.97$\pm$0.03}& 0.97$\pm$0.03& 0.96$\pm$0.07\\
			\textbf{Ex}  & 0.94$\pm$0.04& 0.89$\pm$0.20& 0.94$\pm$0.04& 0.89$\pm$0.20& 0.95$\pm$0.02& 0.95$\pm$0.06& 0.97$\pm$0.02& \textbf{0.98$\pm$0.05}& 0.95$\pm$0.03& 0.97$\pm$0.06&0.98$\pm$0.02& \textbf{0.98$\pm$0.05}\\
			\textbf{EC}  &  0.94$\pm$0.04& 0.89$\pm$0.13& 0.94$\pm$0.04& 0.89$\pm$0.22& 0.95$\pm$0.03& 0.94$\pm$0.08& 0.96$\pm$0.03& \textbf{0.96$\pm$0.08}& 0.96$\pm$0.03& \textbf{0.96$\pm$0.06}& 0.97$\pm$0.03& \textbf{0.96$\pm$0.08}\\
	        \textbf{Fr}  & 0.95$\pm$0.03& 0.93$\pm$0.10& 0.95$\pm$0.03& 0.93$\pm$0.10& 0.96$\pm$0.03& 0.96$\pm$0.06& 0.97$\pm$0.02&\textbf{ 0.98$\pm$0.03}& 0.98$\pm$0.01& \textbf{0.98$\pm$0.02}& 0.98$\pm$0.03& 0.97$\pm$0.05\\
	\hline
   \end{tabular}
    \begin{tabular}{|l|cc|cc|cc|cc|cc|cc|}
   & \multicolumn{12}{c|}{\textbf{Bimodal}}	\\ \cline{2-13}
     \bf Trait & \multicolumn{2}{c|}{\textbf{Kin + AU FF}}  & \multicolumn{2}{c|}{\textbf{Kin + AU DF}} & \multicolumn{2}{c|}{\textbf{Kin + Aud  FF}} & \multicolumn{2}{c|}{\textbf{Kin + Aud  DF}}  & \multicolumn{2}{c|}{\textbf{AU + Aud FF}}  & \multicolumn{2}{c|}{\textbf{AU + Aud  DF}}\\
		& \textbf{Acc} & \textbf{PCC}  & \textbf{Acc} & \textbf{PCC} & \textbf{Acc} & \textbf{PCC} & \textbf{Acc} & \textbf{PCC} & \textbf{Acc} & \textbf{PCC} & \textbf{Acc} & \textbf{PCC}\\
     \hline
      \textbf{Ov}   & 0.97$\pm$0.03& 0.93$\pm$0.15& 0.95$\pm$0.04& 0.89$\pm$0.21& 0.97$\pm$0.02& 0.96$\pm$0.07& 0.96$\pm$0.03& 0.95$\pm$0.06& 0.97$\pm$0.03&\textbf{ 0.97$\pm$0.07}& 0.96$\pm$0.03& 0.95$\pm$0.09\\
			\textbf{RH}    & 0.96$\pm$0.04& 0.92$\pm$0.16& 0.94$\pm$0.04& 0.90$\pm$0.19& 0.97$\pm$0.03& \textbf{0.96$\pm$0.07}& 0.96$\pm$0.03& 0.95$\pm$0.07& 0.97$\pm$0.03& \textbf{0.96$\pm$0.07}& 0.96$\pm$0.03& 0.95$\pm$0.08\\
			\textbf{Ex}  & 0.96$\pm$0.04& 0.93$\pm$0.15& 0.93$\pm$0.04& 0.91$\pm$0.14& 0.97$\pm$0.02& \textbf{0.98$\pm$0.04}& 0.96$\pm$0.02& 0.97$\pm$0.05& 0.97$\pm$0.02& \textbf{0.98$\pm$0.04}& 0.97$\pm$0.02& 0.97$\pm$0.05\\
			\textbf{EC}  & 0.96$\pm$0.04& 0.94$\pm$0.13& 0.95$\pm$0.04& 0.91$\pm$0.16& 0.97$\pm$0.03& 0.95$\pm$0.07& 0.95$\pm$0.03& 0.94$\pm$0.10& 0.97$\pm$0.02& \textbf{0.96$\pm$0.06}& 0.96$\pm$0.03& 0.95$\pm$0.09\\
			\textbf{Fr}  & 0.97$\pm$0.03& 0.96$\pm$0.08& 0.95$\pm$0.03& 0.94$\pm$0.09& 0.97$\pm$0.03& 0.97$\pm$0.07& 0.96$\pm$0.03& 0.96$\pm$0.05& 0.97$\pm$0.02&\textbf{ 0.98$\pm$0.04}& 0.96$\pm$0.03& 0.96$\pm$0.05\\
	    \hline
    \end{tabular}
  \label{tab:MIT_reg}
    \centering
    \fontsize{7}{7}\selectfont
    \caption{Unimodal and multimodal regression results on the FICS dataset. Accuracy and PCC values for different methods are tabulated, with highest PCC achieved per trait denoted in \textbf{bold}.} \vspace{-2mm}
    \begin{tabular}{|l|cc|cc|cc|cc|cc|cc|}
    \hline
    \bf  & \multicolumn{6}{c|}{\textbf{Unimodal}} & \multicolumn{6}{c|}{\textbf{Trimodal}} \\ \cline{2-13}
     \bf Trait & \multicolumn{2}{c|}{\textbf{LSTM Kin}}  & \multicolumn{2}{c|}{\textbf{LSTM AU}} & \multicolumn{2}{c|}{\textbf{LSTM Audio}} & \multicolumn{2}{c|}{\textbf{LSTM FF}}  & \multicolumn{2}{c|}{\textbf{LSTM DF}}  & \multicolumn{2}{c|}{\textbf{LSTM AF}}\\
		& \textbf{Acc} & \textbf{PCC}  & \textbf{Acc} & \textbf{PCC} & \textbf{Acc} & \textbf{PCC} & \textbf{Acc} & \textbf{PCC} & \textbf{Acc} & \textbf{PCC} & \textbf{Acc} & \textbf{PCC}\\
     \hline
      \textbf{Open} & 0.872 & 0.060 & 0.889	& 0.370 & 0.896	& 0.436	& 0.895	& 0.464	& 0.900	& \textbf{0.501} & 0.895 & 0.483\\
			\textbf{Con} & 0.864 & 0.027 & 0.882 & 0.317 & 0.888 & 0.418 & 0.891 & 0.434 & 0.894 & \textbf{0.504} & 0.888 & 0.428\\
			\textbf{Extra} & 0.869 & 0.048	& 0.891	& 0.491	& 0.895	& 0.445	& 0.894	& 0.540 & 0.900 &\textbf{ 0.566} & 0.896 & 0.534\\
			\textbf{Agree} & 0.885 & 0.046	& 0.897	& 0.251	& 0.894	& 0.291	& 0.895	& 0.304	& 0.901	& \textbf{0.377} & 0.899 & 0.300\\
			\textbf{Neuro} & 0.867 & 0.051	& 0.885	& 0.370 & 0.887	& 0.465	& 0.890 & 0.484	& 0.895	& \textbf{0.517} & 0.891 & 0.481\\
      \hline
    \end{tabular}
        \begin{tabular}{|l|cc|cc|cc|cc|cc|cc|}
     & \multicolumn{12}{c|}{\textbf{Bimodal}}	\\ \cline{2-13}
     \bf Trait & \multicolumn{2}{c|}{\textbf{Kin + AU FF}}  & \multicolumn{2}{c|}{\textbf{Kin + AU DF}} & \multicolumn{2}{c|}{\textbf{Kin + Aud  FF}} & \multicolumn{2}{c|}{\textbf{Kin + Aud  DF}}  & \multicolumn{2}{c|}{\textbf{AU + Aud FF}}  & \multicolumn{2}{c|}{\textbf{AU + Aud  DF}}\\
		& \textbf{Acc} & \textbf{PCC}  & \textbf{Acc} & \textbf{PCC} & \textbf{Acc} & \textbf{PCC} & \textbf{Acc} & \textbf{PCC} & \textbf{Acc} & \textbf{PCC} & \textbf{Acc} & \textbf{PCC}\\
     \hline
      \textbf{Open} & 0.892	& 0.368	& 0.893	& 0.382	& 0.892	& 0.418	& 0.898	& 0.456	& 0.893	& 0.459	& 0.898	& \textbf{0.484}\\
			\textbf{Con} & 0.880 & 0.304 & 0.881 & 0.282 & 0.883 & 0.387 & 0.890 & 0.446 & 0.887 & 0.450 & 0.895 & \textbf{0.510}\\
			\textbf{Extra} & 0.893 & 0.474	& 0.891	& 0.485	& 0.888	& 0.415	& 0.892	& 0.450 & 0.895	& 0.531	& 0.900 & \textbf{0.550}\\
			\textbf{Agree} & 0.892 & 0.253	& 0.896 & 0.275	& 0.891	& 0.265	& 0.896	& 0.324	& 0.895	& 0.311	& 0.900 & \textbf{0.378}\\
			\textbf{Neuro} & 0.884 & 0.365	& 0.887	& 0.387	& 0.880 & 0.410 & 0.890	& 0.472	& 0.888	& 0.455	& 0.897	& \textbf{0.533}\\
			
      \hline
    \end{tabular}
    \label{tab:FICS_reg}
    \vspace{3mm}


    \centering
	\fontsize{7}{7}\selectfont
	\renewcommand{\arraystretch}{1.5}
    \caption{Unimodal and multimodal classification results on the MIT dataset. Accuracy and F1-score  are tabulated, with highest F1 achieved per trait denoted in \textbf{bold}.} \vspace{-2mm}
    \begin{tabular}{|l|cc|cc|cc|cc|cc|cc|}
    \hline
    & \multicolumn{6}{c|}{\textbf{Unimodal}} & \multicolumn{6}{c|}{\textbf{Trimodal}} \\ \cline{2-13}
     \bf Trait & \multicolumn{2}{c|}{\textbf{LSTM Kin}}  & \multicolumn{2}{c|}{\textbf{LSTM AU}} & \multicolumn{2}{c|}{\textbf{LSTM Audio}} & \multicolumn{2}{c|}{\textbf{LSTM FF}}  & \multicolumn{2}{c|}{\textbf{LSTM DF}}  & \multicolumn{2}{c|}{\textbf{LSTM AF}}\\
		& \textbf{Acc} & \textbf{F1}  & \textbf{Acc} & \textbf{F1} & \textbf{Acc} & \textbf{F1} & \textbf{Acc} & \textbf{F1} & \textbf{Acc} & \textbf{F1} & \textbf{Acc} & \textbf{F1}\\
     \hline
      \textbf{Ov} & 0.83$\pm$0.11 & 0.82$\pm$0.13 & 0.82$\pm$0.14 & 0.81$\pm$0.15 & 0.94$\pm$0.08 & 0.93$\pm$0.10  & 0.97$\pm$0.07 & 0.96$\pm$0.10 & 0.97$\pm$0.05 &\textbf{0.97$\pm$0.06} & 0.97$\pm$0.06&\textbf{0.97$\pm$0.07}\\
			\textbf{RH}    &  0.79$\pm$0.12 & 0.79$\pm$0.12 & 0.82$\pm$0.12 & 0.82$\pm$0.13 & 0.95$\pm$0.07 & 0.95$\pm$0.07 & 0.95$\pm$0.09 & 0.95$\pm$0.10 &0.98$\pm$0.06 & \textbf{0.98$\pm$0.06}& 0.95$\pm$0.08 & 0.95$\pm$0.10\\
			\textbf{Ex}  &  0.82$\pm$0.13 & 0.82$\pm$0.13 & 0.83$\pm$0.13 & 0.83$\pm$0.14 & 0.95$\pm$0.07 & 0.95$\pm$0.08 & 0.97$\pm$0.06 & 0.96$\pm$0.06 & 0.97$\pm$0.04 & \textbf{0.97$\pm$0.05}&0.96$\pm$0.06 & 0.96$\pm$0.06\\
			\textbf{EC}  &  0.79$\pm$0.13  & 0.79$\pm$0.13 & 0.81$\pm$0.12 & 0.80$\pm$0.13 & 0.93$\pm$0.08 & 0.91$\pm$0.10 & 0.95$\pm$0.07 & 0.94$\pm$0.10 & 0.95$\pm$0.07 &\textbf{0.95$\pm$0.08}&0.94$\pm$0.09 & 0.93$\pm$0.10\\
			\textbf{Fr}  &  0.80$\pm$0.15 & 0.80$\pm$0.16 & 0.86$\pm$0.09 & 0.85$\pm$0.09 & 0.94$\pm$0.07 & 0.94$\pm$0.08 & 0.96$\pm$0.06 & 0.95$\pm$0.06 & 0.97$\pm$0.05 &\textbf{0.96$\pm$0.05}&0.95$\pm$0.06 & 0.94$\pm$0.07\\
	\hline
    \end{tabular}
    \begin{tabular}{|l|cc|cc|cc|cc|cc|cc|}
     & \multicolumn{12}{c|}{\textbf{Bimodal}}	\\ \cline{2-13}
    \bf Trait & \multicolumn{2}{c|}{\textbf{Kin + AU FF}}  & \multicolumn{2}{c|}{\textbf{Kin  + AU DF}} & \multicolumn{2}{c|}{\textbf{Kin + Aud  FF}} & \multicolumn{2}{c|}{\textbf{Kin + Aud  DF}}  & \multicolumn{2}{c|}{\textbf{AU + Aud FF}}  & \multicolumn{2}{c|}{\textbf{AU + Aud DF}}\\
		& \textbf{Acc} & \textbf{F1}  & \textbf{Acc} & \textbf{F1} & \textbf{Acc} & \textbf{F1} & \textbf{Acc} & \textbf{F1} & \textbf{Acc} & \textbf{F1} & \textbf{Acc} & \textbf{F1}\\
     \hline
      \textbf{Ov}   & 0.80$\pm$0.14 & 0.80$\pm$0.14 & 0.85$\pm$0.13 & 0.85$\pm$0.14 & 0.96$\pm$0.08 & \textbf{0.96$\pm$0.09} & 0.96$\pm$0.07 & \textbf{0.96$\pm$0.08} & 0.97$\pm$0.07 & \textbf{0.96$\pm$0.07} & 0.96$\pm$0.06 &  \textbf{0.96$\pm$0.07}\\
			\textbf{RH}    & 0.79$\pm$0.12 & 0.79$\pm$0.13 & 0.83$\pm$0.13 & 0.83$\pm$0.14 & 0.95$\pm$0.08 & 0.94$\pm$0.10 & 0.96$\pm$0.07 & \textbf{0.96$\pm$0.08} & 0.95$\pm$0.07 & 0.95$\pm$0.08 & 0.95$\pm$0.07 & 0.95$\pm$0.08\\
			\textbf{Ex}  & 0.81$\pm$0.12 & 0.80$\pm$0.12 & 0.84$\pm$0.11 & 0.83$\pm$0.12 & 0.95$\pm$0.06 & 0.95$\pm$0.07 & 0.96$\pm$0.07 & 0.95$\pm$0.08 & 0.97$\pm$0.05 & \textbf{0.97$\pm$0.05} & 0.97$\pm$0.05 & 0.96$\pm$0.05\\
			\textbf{EC}  & 0.78$\pm$0.12 & 0.76$\pm$0.13 & 0.84$\pm$0.13 & 0.83$\pm$0.14 & 0.93$\pm$0.09 & 0.92$\pm$0.12 & 0.94$\pm$0.08 & \textbf{0.94$\pm$0.08 }& 0.93$\pm$0.07 & 0.92$\pm$0.08 & 0.94$\pm$0.08 & 0.93$\pm$0.10\\
			\textbf{Fr}  & 0.84$\pm$0.10 & 0.84$\pm$0.11 & 0.87$\pm$0.11 & 0.86$\pm$0.12 & 0.95$\pm$0.06 & 0.94$\pm$0.07 & 0.96$\pm$0.07 & 0.95$\pm$0.07 & 0.95$\pm$0.07 & 0.95$\pm$0.07 & 0.97$\pm$0.06 & \textbf{0.96$\pm$0.06}\\
			
      \hline
    \end{tabular}
    \label{tab:MIT_cls}\vspace{-2mm}
\end{table*}

\begin{table*}[!htbp]
    \centering
    \fontsize{7}{7}\selectfont
    \caption{Unimodal and multimodal classification results on the FICS dataset. Accuracy and F1-score  for different methods are tabulated, with highest F1 achieved per trait denoted in \textbf{bold}.} \vspace{-2mm}
    \begin{tabular}{|l|cc|cc|cc|cc|cc|cc|}
    \hline
     & \multicolumn{6}{c|}{\textbf{Unimodal}} & \multicolumn{6}{c|}{\textbf{Trimodal}} \\ \cline{2-13}
     \bf Trait & \multicolumn{2}{c|}{\textbf{LSTM Kin}}  & \multicolumn{2}{c|}{\textbf{LSTM AU}} & \multicolumn{2}{c|}{\textbf{LSTM Audio}} & \multicolumn{2}{c|}{\textbf{LSTM FF}}  & \multicolumn{2}{c|}{\textbf{LSTM DF}}  & \multicolumn{2}{c|}{\textbf{LSTM AF}}\\
		& \textbf{Acc} & \textbf{F1}  & \textbf{Acc} & \textbf{F1} & \textbf{Acc} & \textbf{F1} & \textbf{Acc} & \textbf{F1} & \textbf{Acc} & \textbf{F1} & \textbf{Acc} & \textbf{F1}\\
     \hline
      \textbf{Open}   &  0.519 & 0.516 & 0.635 & 0.634 & 0.595 & 0.590 & 0.638 & 0.638 & 0.676 & \textbf{0.672} &0.633 & 0.633\\
			\textbf{Con}    &  0.513 & 0.513 & 0.618 & 0.618 & 0.599 & 0.592 & 0.623 & 0.623 & 0.640 & \textbf{0.638} &0.594 & 0.582\\
			\textbf{Extra}  &  0.505 & 0.505 & 0.651 & 0.651 & 0.624 & 0.623 & 0.667 & 0.665 & 0.695 & \textbf{0.695} &0.671 & 0.669\\
			\textbf{Agree}  &  0.481 & 0.479 & 0.580 & 0.580 & 0.551 & 0.545 & 0.588 & 0.585 & 0.599 & \textbf{0.598} &0.565 & 0.560\\
			\textbf{Neuro}  &  0.523 & 0.518 & 0.627 & 0.624 & 0.578 & 0.547 & 0.618 & 0.611 & 0.665 & \textbf{0.665} &0.639& 0.638\\
    \end{tabular}
    \centering
 \fontsize{7}{7}\selectfont
    \begin{tabular}{|l|cc|cc|cc|cc|cc|cc|}
    \hline
     & \multicolumn{12}{c|}{\textbf{Bimodal}}	\\ \cline{2-13}
     \bf Trait & \multicolumn{2}{c|}{\textbf{Kin + AU FF}}  & \multicolumn{2}{c|}{\textbf{Kin + AU DF}} & \multicolumn{2}{c|}{\textbf{Kin + Aud FF}} & \multicolumn{2}{c|}{\textbf{Kin + Aud DF}}  & \multicolumn{2}{c|}{\textbf{AU + Aud FF}}  & \multicolumn{2}{c|}{\textbf{AU + Aud DF}}\\
		& \textbf{Acc} & \textbf{F1}  & \textbf{Acc} & \textbf{F1} & \textbf{Acc} & \textbf{F1} & \textbf{Acc} & \textbf{F1} & \textbf{Acc} & \textbf{F1} & \textbf{Acc} & \textbf{F1}\\
     \hline
      \textbf{Open}&0.629 & 0.628 & 0.632 & 0.632 & 0.612 & 0.598 & 0.641 & 0.638 & 0.667 & 0.664 &0.677 &\textbf{ 0.672}\\
			\textbf{Con}&0.604 & 0.604 & 0.599 & 0.598 & 0.564 & 0.564 & 0.609 & 0.607 & 0.620 & 0.619 &0.637 & \textbf{0.637}\\
			\textbf{Extra}&0.648 & 0.648 & 0.657 & 0.653 & 0.649 & 0.649 & 0.639 & 0.636 & 0.668 & 0.667 &0.682 & \textbf{0.682}\\
			\textbf{Agree}&0.593 & 0.586 & 0.584 & 0.583 & 0.544 & 0.450 & 0.592 & 0.592 & 0.601 & 0.594 &0.603 & \textbf{0.596}\\
			\textbf{Neuro}&0.626 & 0.623 & 0.620 & 0.616 & 0.617 & 0.615 & 0.637 & 0.635 & 0.652 & \textbf{0.651 }&0.656& \textbf{0.651}\\
			
      \hline
    \end{tabular}
    \label{tab:FICS_cls}\vspace{2mm}
\centering
\fontsize{7}{7}\selectfont
\caption{Soft Additive Attention Fusion Results over the 2s behavioral slice: MIT Dataset (left) and FICS Dataset (right)}\vspace{-2mm}
\label{tab:AF_FICS}   
 
\begin{tabular}{|l|cc |cc|} 
\hline
     \bf Trait & \multicolumn{2}{c|}{\textbf{Classification}} & \multicolumn{2}{c|}{\textbf{Regression}}    \\ 
               & \textbf{Acc} & \textbf{F1} & \textbf{Acc} & \textbf{PCC} \\ \hline

\textbf{Ov}    & 0.91±0.09            & 0.89±0.10         & 0.93±0.03            & 0.92±0.09         \\ 
\textbf{RH}    & 091±0.12           & 0.90±0.12          & 0.92±0.03            & 0.92±0.08        \\ 
\textbf{Ex}    & 0.92±0.09           & \textbf{0.91±0.10} & 0.93±0.02             & \textbf{0.94±0.08}        \\ 
\textbf{EC}    & 0.84±0.12            & 0.82±0.14         & 0.91±0.02            & 0.90±0.10        \\ 
\textbf{Fr}    & 0.92±0.10            & \textbf{0.91±0.10} & 0.93±0.02            & \textbf{0.94±0.05} \\
\hline
\end{tabular} 
\vspace*{5mm}
\hspace*{5mm}
\begin{tabular}{|l|cc |cc|} 
\hline
     \bf Trait   & \multicolumn{2}{c|}{\textbf{Classification}} & \multicolumn{2}{c|}{\textbf{Regression}}    \\ 
                & \textbf{Acc} & \textbf{F1} & \textbf{Acc} & \textbf{PCC}\\ \hline

\textbf{O}     & 0.632                  & 0.619               & 0.896                  & 0.475              \\ 
\textbf{C}     & 0.605                  & 0.604              & 0.888                  & 0.428              \\ 
\textbf{E}     & 0.656                  & \textbf{0.656}               & 0.893                  & \textbf{0.501}              \\ 
\textbf{A}     & 0.561                  & 0.556               & 0.899                  & 0.326              \\ 
\textbf{N}     & 0.625                  & 0.625               & 0.887                  & 0.479               \\
\hline
\end{tabular}\vspace{-4mm}
\end{table*}

			

Based on Tables~\ref{tab:MIT_reg}--\ref{tab:FICS_cls}, we make the following observations:
\begin{itemize}
    \item For \textbf{regression} benchmarking (Tables~\ref{tab:MIT_reg},~\ref{tab:FICS_reg}), PCC is a more stringent measure than Acc, as very low PCC values are observed with relatively high Acc values for the FICS dataset (Table~\ref{tab:FICS_reg}). Tables~\ref{tab:MIT_reg} and~\ref{tab:MIT_cls} show that regression and classification results are comparable for the (smaller) MIT dataset. For FICS, the regression scores are considerably higher than the classification scores, which can be attributed to Gaussian-distributed FICS traits with means around 0.5~\cite{escalante2020modeling}.
    
    \item Speech features achieve optimal interview trait prediction~(Table~\ref{tab:MIT_reg}), while Kineme and AU features perform comparably. Optimal personality trait regression is also achieved with audio features (Table~\ref{tab:FICS_reg}), even as AUs significantly outperform kinemes on the FICS dataset.
    
    \item Higher PCC scores are achieved with multimodal as compared to unimodal methods on both the MIT and FICS datasets. Bimodal and trimodal fusion perform very similarly for both interview and personality trait prediction, with maximum PCC values of 0.98 achieved for the Excited and Friendliness interview traits, and a peak PCC of 0.566 achieved for the Extraversion personality trait on FICS obtained with trimodal fusion. 
    
    \item Focusing on multimodal methods, bimodal combinations involving audio outperform others for interview trait prediction. Also, feature fusion is more effective than decision fusion in this case. Slightly different trends are noted for the FICS dataset with decision fusion slightly outperforming feature fusion; optimal PCC values are noted for the AU$+$Aud combination with decision fusion, implying that speech features individually and in combination with others acquire high predictive power, mirroring findings in~\cite{naim2016automated}. Bimodal predictions improving over unimodal ones conveys that kinemes and AUs provide complementary information concerning interview and personality traits.
    
    \item Among trimodal fusion methods, decision fusion slightly outperforms attention and feature fusion on the MIT dataset, while decision, attention and feature fusion approaches perform first, second and third best on the FICS dataset. These results again reveal the complementary utility of the kineme, AU and speech features; optimal performance achieved with trimodal decision fusion conveys that the AU and kineme classifiers improve prediction performance in instances where speech descriptors are  ineffective.
    
    \item Focusing on \textbf{classification} (Tables~\ref{tab:MIT_cls} and~\ref{tab:FICS_cls}), considering unimodal results, audio features achieve optimal F1-scores on Interview traits (highest F1 of 0.95 for Recommended Hiring and Excited), while AUs achieve the best classification on personality traits (maximum F1 of 0.651 for Extraversion). AUs and kinemes perform similarly on the MIT dataset, while speech descriptors achieve much higher F1-scores than kinemes on FICS.  
  
  \item Multimodal approaches again outperform unimodal methods in categorizing both interview and personality traits. With respect to bimodal methods, combinations involving speech tend to perform well for both interview and personality prediction. There is little to choose between feature and decision fusion for interview trait prediction, while decision fusion slightly outperforms feature fusion for predicting personality traits. 
  
  \item Trimodal fusion performs best producing peak F1 scores of 0.98 and 0.695 for the RH interview, and Extraversion personality traits. Decision fusion produces optimal trait classification on both datasets, with feature and attention fusion performing comparably.
\end{itemize}
The above results represent trait prediction at the~\textbf{video level}, on examining 15s FICS videos or upon collating classification/regression results over 5--60s chunks/segments on the MIT dataset (the best results obtained by averaging chunk-level values, or computing the majority label over all chunks are listed in Tables~\ref{tab:MIT_reg} and~\ref{tab:MIT_cls}).
 
   
   
\subsubsection{Thin-slice predictions:}  
We explore trait prediction over short behavioral episodes known as \textit{thin slices} and present the multimodal results for classification and regression using soft additive attention-fusion over 2s behavioral slice in Table~\ref{tab:AF_FICS}. The results convey that reasonable prediction performance can be achieved even with 2s-long slices expressing the efficacy of these small behavioral slices for predicting different traits. \\
Further, we visualize the comparison of chunk and video-level prediction performance for varying time-lengths over all three modalities in Figures~\ref{fig:Chunk_vs_Vid_kineme}--\ref{fig:Chunk_vs_Vid_Audio}. It can be noted that better prediction performance has been achieved with video-level as compared to chunk-level implying that while episodic behaviors may be inconsistent with one another, trait specific behaviors tend to be homogeneous over longer time-span. For the OCEAN traits, kineme-based chunk and video-level PCC values deteriorate over larger time-slices (Fig.~\ref{fig:Chunk_vs_Vid_kineme} (left)) while remaining stable in the case of AU features. The speech features are mostly consistent with different time-slices in case of chunk-level while decreasing slightly for video-level prediction. Conversely, chunk and video-level PCC values increase for all three modalities with increasing time-slice length for the MIT dataset (Fig.~\ref{fig:Chunk_vs_Vid_kineme}--\ref{fig:Chunk_vs_Vid_Audio} (right)). This trend highlights that AUs, describing facial behavior, encode more trait specific information, specifically for personality traits as compared to kinemes characterizing head movement and speech features. The better prediction with larger time-slices for all modalities over the MIT dataset suggests that interview behavior can be captured better over longer time-span.

\section{Explainability \& Interpretability}

\subsection{Interpretation via kinemes and AUs}
Along with their predictive power, kinemes and AUs also enable facile trait-specific behavioral explanations. To this end, we considered the top and bottom 10-percentile videos for each trait, and computed the most frequently occurring AUs and kinemes for the same. The most frequently occurring four kinemes, and five dominant AUs for these high (H) and low (L)-rated videos are presented in Table~\ref{tab:explainableKineme}. Analysing the table, we make the following remarks:

\begin{table*}[!htbp]
    \centering
    \fontsize{7}{7}
    \caption{Explaining OCEAN and interview traits via kinemes and AUs. MIT kinemes in bold font are visualized in Figure~\ref{fig:selectKinemes_MIT}.} \vspace{-3mm}
    \begin{tabular}{c|l|c|c|l}
    \toprule
     \bf Dataset & \bf Trait  & \bf Dominant Kin & \bf Dominant AUs & \bf Inferences \\
     \hline
      \multirow{10}{*}{\textbf{FICS}} & \textbf{O (H)} & 2, 8, 10, 16 &  7, 12, 14, 25, 26 & Persistent head movements (as noted in~\cite{KOPPENSTEINER2013}) with nodding and smiling.\\
      &\textbf{C (H)} & 1, 8, 10, 16 & 7, 12, 17, 25, 26 & Upward head-tilt indicative of upright demeanor and head nodding. \\
      &\textbf{E (H)} & 2, 10, 14, 16 & 10, 12, 17, 25, 26 & Head tilt-down with nodding, and facial gestures related to speaking.\\
      &\textbf{A (H)} & 3, 8, 10, 16 & 7, 12, 14, 25, 26 & Frequent head nodding and smiling (associated with courteous behavior~\cite{Ishii2020,Kawahara18}).\\ 
      &\textbf{N (H)} & 2, 8, 10, 16 & 7, 12, 17, 25, 26 & Frequent head movements with nodding and smiling.\\ 
			&\textbf{O (L)} & 1, 6, 11, 16 & 4, 10, 14, 17, 26 & Relatively fewer head movements and frowning. \\
			&\textbf{C (L)} & 2, 4, 8, 16 & 4, 7, 10, 14, 25 & Head tilt-down avoiding eye-contact, head shaking and frowning.\\
			&\textbf{E (L)} & 1, 4, 10, 16 & 4, 7, 10, 14, 17 & Tilt-up, head shaking and frowning.  \\ 
			&\textbf{A (L)} & 1, 8, 9, 16 & 4, 14, 17, 25, 26 & Frequent head movements and frowning.\\
			&\textbf{N (L)} & 1, 5, 12, 16 & 4, 7, 10, 14, 25 & Few head movements, head shaking and frowning.\\ \hline
			 \multirow{8}{*}{\textbf{MIT}} & \textbf{RH (H)} & \textbf{16}, \textbf{14}, 3, 4  & 5, 10, 12, 14, 25 & Head nodding and smiling, and being expressive.\\
      &\textbf{Ex (H)} & \textbf{14}, 3, 4, 9   & 5, 10, 12, 14, 25 & Head nodding and exhibiting persistent head motion. Smiling and expressive.\\
      &\textbf{EC (H)} & \textbf{14}, 12, 4, 5  & 6, 7, 10, 14, 25   & Head up, nodding and showing limited facial emotions.\\
      &\textbf{Fr (H)} & \textbf{16}, 3, \textbf{11}, \textbf{14} & 5, 10, 12, 14, 25 &  Frequent head movements and smiling.\\ 
			&\textbf{RH (L)} & \textbf{11}, 1, 2, 5   & 6, 7, 12, 14, 25  & Head shaking and exhibiting minimal facial expressions.\\
      &\textbf{Ex (L)} & \textbf{11}, \textbf{16}, 2, 3  & 4, 6, 7, 14, 25   &  Head shaking and nodding. Frowning and showing minimal facial expressions. \\
      &\textbf{EC (L)} & 13, 7, \textbf{16}, \textbf{11} & 6, 7, 10, 12, 25   & Frequent nodding is perceived as avoiding eye-contact.\\
      &\textbf{Fr (L)} &  3, \textbf{11}, 4, 9  & 1, 4, 6, 7, 25 & Head shaking, frowning and otherwise being minimally expressive.\\ 
      \bottomrule
      \hline
    \end{tabular}
    \label{tab:explainableKineme}\vspace{-4mm}
\end{table*}
\begin{itemize}
    \item The presence of kineme 16 (denoting head nodding and shaking) in all OCEAN traits conveys the significance of head motion for the characterization of personality traits. Combination of head nodding and shaking with other kineme representations highlights the subtle difference between high and low-rated personality impressions. Also, note that AUs 25 and 26 signifying talking behavior are present in all videos.
    \item Focusing on other kinemes, high Openness is characterized by kinemes 2 and 8, which signify persistent head movements. This finding is echoed in~\cite{KOPPENSTEINER2013}, where large motion variations are found to associate with high O impressions. Presence of AUs 12 and 14 indicates that a smiling demeanor characterizes high O. Conversely, kineme 6 denoting minimal head motion and AUs 4 and 17 typical of frowning and diffident behavior are commonly noted for low O videos.
\item Kineme 1 denoting an upward head tilt is associated with high C, while kinemes 2 and 4 depicting tilt-down and head-shaking are associated with low C. This indicates that attempting to maintain eye-contact conveys diligence and honesty, while avoiding eye-contact conveys insincerity. 
\item Extraversion appears to be conveyed better by AUs than kinemes; Dominant AUs for high E include 10, 12 and 17 indicating a friendly and talkative nature, while dominant kinemes 2 and 14 convey significant head movements. Conversely, low E is associated with kineme 4 denoting head-shaking and AUs 4, 7 and 17 indicating frowning, overall conveying a socially distant nature.
\item High Agreeableness is characterized by kineme 3 (head-nod), and AUs 12 and 14 which constitute a smile. Conversely, kinemes 1, 8 and 9 dominate low A, and they collectively convey persistent head motion. Also, AUs dominant for low A are 4, 14 and 17, cumulatively describing a frown; overall, nodding and smiling is viewed as courteous, while frequent head movements and frowning convey hostility.  
\item Emotional stability (high N) is associated with kinemes 2 and 8, and AUs 7, 12 and 17, indicating persistent head motion and facial expressiveness. On the other hand, a neurotic trait is conveyed via limited head motion and head-shaking (kinemes 1, 5, 12) and frowning (described by AUs 4, 7, 10).
\item While kinemes for the MIT videos are less discernible, due to smaller face size (Fig.~\ref{fig:dataset_frames}) and the fact that they capture an interactional setting, some patterns are nevertheless evident as seen in Fig.~\ref{fig:selectKinemes_MIT}; these kinemes are highlighted in Table~\ref{tab:explainableKineme}. As with FICS, Kineme 14 denoting a head-nod is commonly observed for all high trait videos, while kineme 11 depicting a head-shake is common for all low-trait videos.  
\item High RH scores are elicited with expressive facial behavior involving head-nodding and smiling.   Conversely, low RH scores are associated with head-shaking and exhibiting limited facial expressions. Highly excited behavior is associated with identical AUs as high RH, and persistent head motion. Inversely, low excitement scores are connected with head shaking, and limited facial emotions. 
\item Identical AUs are observed for both high and low eye-contact, implying that head movements primarily impact eye-contact impressions. Head nodding (kineme 14) is associated with high EC, while kinemes 11 and 16 depicting head shaking and frequent head-nodding elicit low EC scores. Therefore interestingly, while head nodding is beneficial, frequent nodding is perceived as avoiding eye-contact. 
\item High friendliness is characterized by kinemes 11, 14 and 16, signifying persistent head motion along with expressive and smiling facial movements (AUs 5, 12 and 14). Conversely, low friendliness is associated with head-shaking (kineme 11) and frowning (AUs 4, 6, 7).                                              
\end{itemize}
\begin{figure*}[!htbp]
\centering
    \includegraphics[width=0.48\linewidth, height=4.2cm]{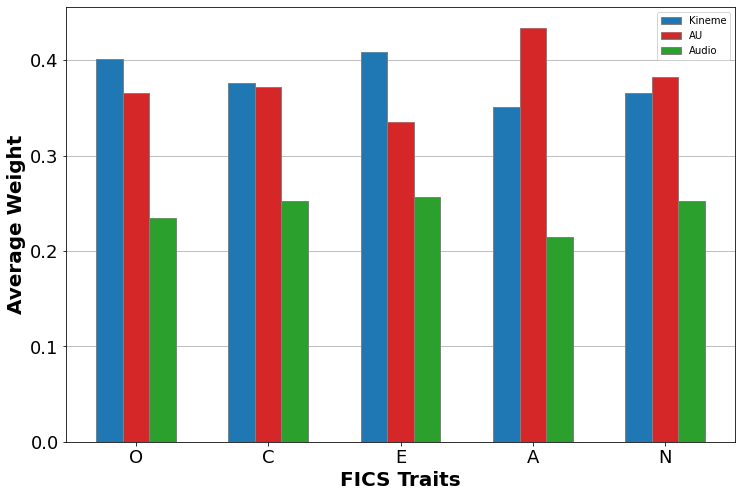}\hspace{0.05cm}\includegraphics[width=0.48\linewidth, height=4.2cm]{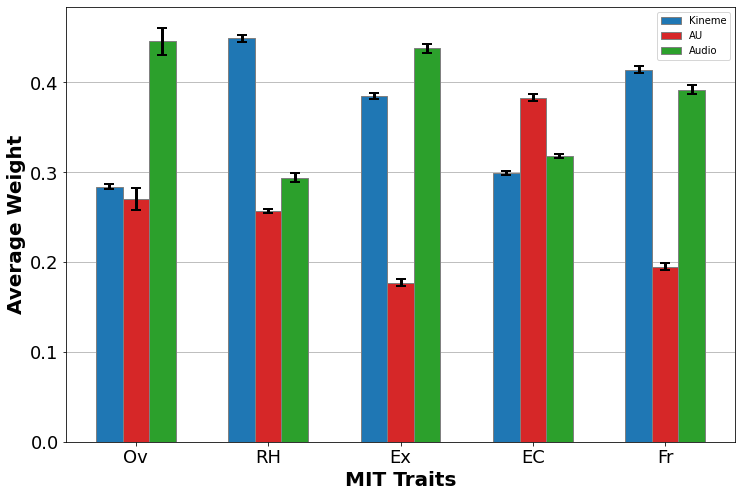}
        \vspace{-4mm}
\caption{Mean modality-specific attention weights for personality traits (left) and interview traits (right). Error bars denote standard error.
  }\label{fig:att_weights}\vspace{-4mm}
\end{figure*}

\subsection{Attention Score-based Interpretations} \label{attention_score}
While Table~\ref{tab:explainableKineme} presents unimodal behavioral explanations via kinemes and AUs, behaviors are expressed and best modeled multimodally as seen from our empirical results (Section~\ref{RandD}). For multimodal explanations, we explore the attention-fusion network (Fig.~\ref{fig:LSTM_attention_arch}) to estimate the relative contribution of each modality towards trait regression. We visualize softmax scores learned by the attention-fusion network as follows. For the \textbf{FICS} dataset, we present mean attention scores obtained over 10 runs on 15s test videos (Fig.~\ref{fig:att_weights}(left)), while we present softmax scores averaged over 15s chunks for \textbf{MIT} videos across 50 runs (Fig.~\ref{fig:att_weights}(right)). Our remarks from the weight plots are as follows:

\begin{itemize}
    \item Cumulatively, Fig.~\ref{fig:att_weights} conveys that while the relative contribution of speech features towards weighted fusion is not high for personality trait prediction, they tend to play a significant role in predicting interview traits on the MIT dataset. These observations mirror prior findings; the criticality of visual features such as head movements and facial movements for personality trait recognition has been noted in~\cite{lepri2012connecting,malik2020empathetic} while the impact of prosodic speech features on interview trait impressions is discussed in~\cite{naim2016automated}. 
    
    \item Fig.~\ref{fig:att_weights}(left) conveys that either kineme or AU features are most critical for personality trait prediction. Specifically, kinemes maximally contribute to the prediction of Openness and Extraversion, while AUs are most critical for predicting Agreeableness and Neuroticism. Both kinemes and AUs are found to be equally critical for estimating Conscientiousness, in line with the findings in \cite{celiktutan2015automatic}. 
   Extraversion and Openness are conveyed by exaggerated physical and head movements~\cite{oberzaucher2008everything,KOPPENSTEINER2013}, with different head movement patterns representing high and low Extraversion~\cite{ruhland2015perception}.  While Agreeableness is also positively correlated with head movements~\cite{oberzaucher2008everything, ruhland2015perception}, empathetic behavior is accurately conveyed via facial expressions as denoted by the higher AU weights. Facial movements (\eg, unconcerned or anxious) better convey emotional stability~\cite{breil202113}. 

\item From Fig.~\ref{fig:att_weights}(right), it can be seen that facial movements have relatively less impact on interview trait prediction with the exception of eye contact. This can partly be attributed to the smaller face size in MIT videos, limiting the efficacy of AU detection. Conversely, speech features significantly impact trait prediction with the exception of recommended hiring and eye contact traits. While prosodic speech behavior has been found to considerably influence interview trait impressions~\cite{degroot2007evidence,naim2016automated}, other forms of non-verbal behavior such as positive facial expressions and frequent postural changes are known to impact hierabilty~\cite{levine2002women}. 

\item For the Excited trait, speech plays a prominent role with a high correlation to continuous or restricted head movement \cite{walther2014less}. On the surprising finding of AUs and speech features impacting eye-contact, prior studies~\cite{eyben2013acoustics} have revealed a low-yet-meaningful correlation between eye contact impressions and vocal acoustic features. Friendliness is best characterized by head movement and voice features, showing that the integration of visual and auditory modalities can be crucial in discerning interviewee friendliness~\cite{house2007integrating}. 
\end{itemize}

\section{Conclusion}
This work demonstrates the efficacy of multimodal (kineme, AU and speech) behavioral cues to achieve explainable prediction of OCEAN and interview traits. Our results confirm that efficient trait prediction can be achieved with both unimodal and multimodal approaches. Also, multimodal approaches outperform their unimodal counterparts owing to complementary information provided by trait-specific behavioral cues. In addition, frequently occurring kineme and AU patterns enable behavioral explanations associated with each trait.

In terms of limitations, this work extracts all behavioral features over a fixed time window (same time-scale); however, behaviors associated with human personality may manifest over different time scales; \eg, facial expression or head motion patterns could be affected by speaking behavior (talkative: drastic variation in speaking behavior over video frames, or reserved: lingering silence over most video frames). Investigating the effect of temporal scales will be a future research direction. Trait-specific behavioral patterns can also be utilized to create virtual agents to train users in interviewing or public speaking settings. The authors do not advise using the proposed methodologies for complex processes like job recruitment \textit{per se}; however, explanatory technologies can be utilized as a complementary tool in decision-making processes.


%



\ifCLASSOPTIONcompsoc
  \section*{Acknowledgments}
\else
  \section*{Acknowledgment}
\fi

We would like to thank A. Samanta, IIT Kanpur for sharing the kineme implementation.

\ifCLASSOPTIONcaptionsoff
  \newpage
\fi



%

%
\bibliographystyle{IEEEtran}
\bibliography{references}
\vskip -3\baselineskip
\begin{IEEEbiography}[{\includegraphics[width=1in,height=1.25in,clip,keepaspectratio]{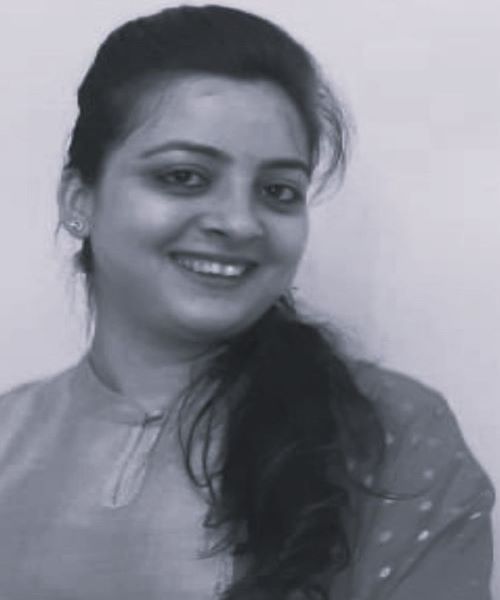}}]{Surbhi Madan}
 is currently pursuing her PhD at the Indian Institute of Technology Ropar, India. She received an M.Tech degree in Computer Science and Engineering from the National Institute of Technology, Hamirpur (India). She received her B.Tech. in Information Technology from Uttar Pradesh Technical University, Lucknow (India). Her research interests include affective computing, computer vision, and Human-Computer Interaction. 
\end{IEEEbiography}
\vskip -2.5\baselineskip plus -1fil
\begin{IEEEbiography}[{\includegraphics[width=1in,height=1.25in,clip,keepaspectratio]{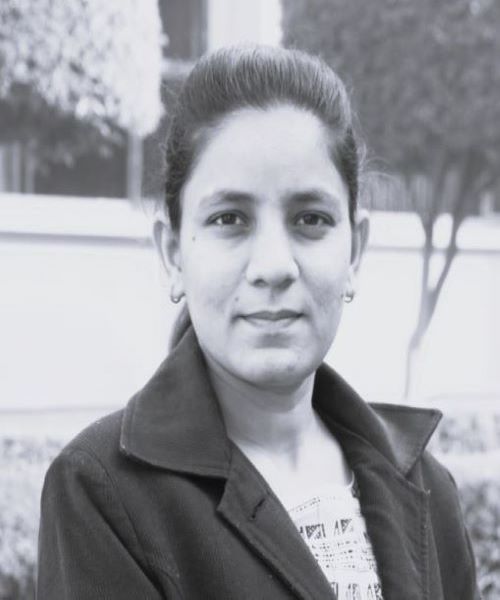}}]{Monika Gahalawat}

received her Master's degree in Computer Science and Engineering from the Indian Institute of Technology Roorkee and Bachelor of Engineering degree from Chandigarh College of Engineering and Technology affiliated with Punjab University in Chandigarh. She's currently a PhD student at the University of Canberra, Australia. Her research interests include computer vision, affective computing, behavior analysis and human-computer interaction.
\end{IEEEbiography}
\vskip -2.5\baselineskip plus -1fil
\begin{IEEEbiography}[{\includegraphics[width=1in,height=1.25in,clip,keepaspectratio]{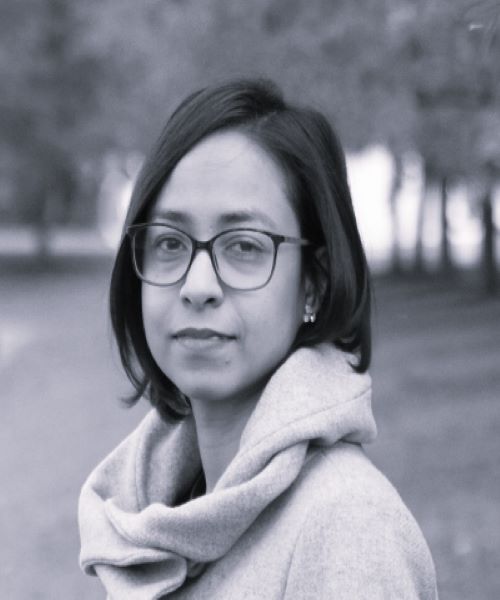}}]{Tanaya Guha}
received her Ph.D.\ degree in Electrical \& Computer Engineering from the University of British Columbia, Vancouver, in 2013. She is a Senior Lecturer of Computing Science at University of Glasgow. Her research focuses on developing machine intelligence capabilities to understand human behavior combining machine learning, speech/signal processing and computer vision. She is an elected member of IEEE MSA TC, IEEE MMSP TC and an executive committee member of AAAC.
\end{IEEEbiography}
\vskip -2.5\baselineskip plus -1fil
\begin{IEEEbiography}[{\includegraphics[width=1in,height=1.25in,keepaspectratio]{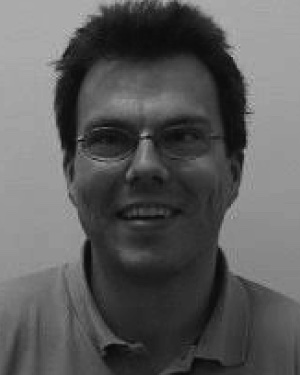}}]{Roland Goecke} received his Ph.D.\ degree in computer science from The Australian National University, Canberra, in 2004. He is Professor of Affective Computing with the University of Canberra, where he is serves as Director of the Human-Centred Technology Research Centre. His research interests include affective computing, pattern recognition, computer vision, human–computer interaction and multimodal signal processing. He is a senior member of the IEEE, and an ACM and AAAC member.
\end{IEEEbiography}
\vskip -2.5\baselineskip plus -1fil
\begin{IEEEbiography}[{\includegraphics[width=1in,height=1.25in,keepaspectratio]{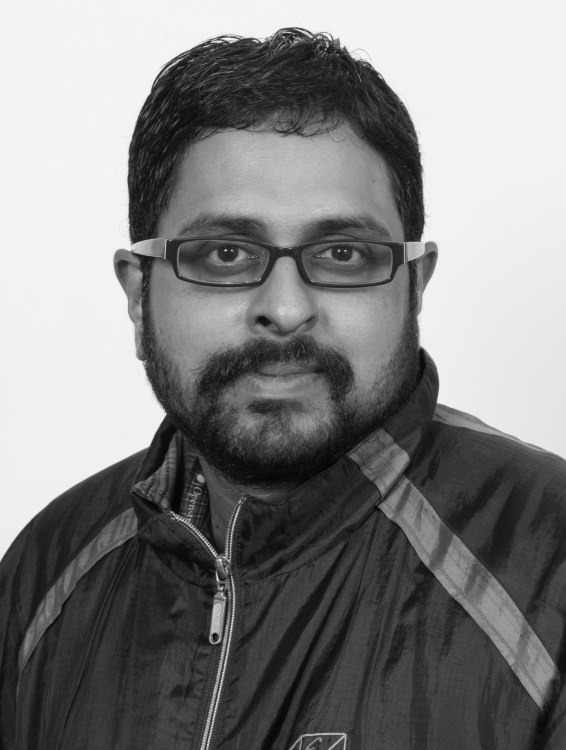}}]{Ramanathan Subramanian} received his Ph.D.\ in Electrical and Computer Engg. from NUS in 2008. He is Associate Professor in the School of IT \& Systems, University of Canberra. His past affiliations include IIT Ropar, IHPC (Singapore), U Glasgow (Singapore), IIIT Hyderabad and UIUC-ADSC (Singapore). His research focuses on Human-centered computing, especially on modeling non-verbal behavioral cues for interactive analytics. He is an IEEE Senior Member, and an ACM and AAAC member.
\end{IEEEbiography}




\end{document}